%% file: colm2025_conference.tex
\documentclass{article} 
\usepackage[preprint]{colm2025_conference}

\title{Skill-Targeted Adaptive Training}

\author{Yinghui He\thanks{Equal contribution.} \quad Abhishek Panigrahi$^*$ \quad Yong Lin \quad Sanjeev Arora  \\
Princeton Language and Intelligence, Princeton University \\
\texttt{\{yh0068, ap34, yl7690, arora\}@princeton.edu} \\
}

\newcommand{\new}

\input{commands.tex}

\definecolor{darkblue}{rgb}{0, 0, 0.5}
\hypersetup{colorlinks=true, citecolor=darkblue, linkcolor=darkblue, urlcolor=darkblue}

\begin{document}

\ifcolmsubmission
\linenumbers
\fi

\maketitle

\input{main_sections/abstract}
\input{main_sections/introduction}

\input{main_sections/preliminary_design}
\input{main_sections/experiments}
\input{main_sections/discussion}

\input{main_sections/related_work}

\input{main_sections/conclusion}
\input{main_sections/ethics_statement}

\bibliography{colm2025_conference}
\bibliographystyle{colm2025_conference}

\newpage
\appendix
\input{appendix/related_work}

\input{appendix/design_details}
\input{appendix/full_experiment_details}
\input{appendix/additional_results}

\input{appendix/ablation_analysis}

\end{document}

%% file: commands.tex
\usepackage{microtype}      
\usepackage{hyperref}       
\usepackage{url}            
\usepackage{booktabs}       
\usepackage{lineno}         
\usepackage{amsmath}        
\usepackage{amssymb}        
\usepackage{amsfonts}       
\usepackage{nicefrac}       
\usepackage{xcolor, pifont} 
\usepackage[table]{xcolor}  
\usepackage{twemojis}       
\usepackage{multirow}       
\usepackage{subcaption}     
\usepackage{wrapfig}        
\usepackage{makecell}       
\usepackage{array}          
\usepackage{caption}        
\usepackage{fontawesome5}   
\usepackage{minitoc}        
\usepackage{graphicx}       
\usepackage{fdsymbol}       
\usepackage[most]{tcolorbox}
\usepackage{cleveref}       
\usepackage{algorithm}      
\usepackage{algpseudocode}  
\usepackage{listings}       
\usepackage{textcomp}       
\definecolor{lightgray}{gray}{0.9} 
\definecolor{myred}{HTML}{D38486}
\definecolor{myblue}{HTML}{CDDFFA}
\definecolor{mydarkblue}{HTML}{3B87E8}
\definecolor{mydarkdarkblue}{HTML}{7689A6}
\definecolor{mypink}{HTML}{F9D9E4}
\definecolor{mylightpink}{HTML}{FBE3EB}
\definecolor{mylightblue}{HTML}{D4E4F9}

\definecolor{sage1}{HTML}{E6F0E7}
\definecolor{sage2}{HTML}{CCDBCB}
\definecolor{sage3}{HTML}{B2C5AF}
\definecolor{sage4}{HTML}{8FA58B}
\definecolor{sage5}{HTML}{748A6F}
\definecolor{sage6}{HTML}{324D3E}

\definecolor{sagetext}{HTML}{5FA163}

\definecolor{green1}{HTML}{DDE6D4}
\definecolor{green2}{HTML}{A3B899}
\definecolor{green3}{HTML}{667B68}
\definecolor{pink1}{HTML}{F6E5E7}
\definecolor{pink2}{HTML}{EFC0BC}
\definecolor{pink3}{HTML}{E69B97}
\definecolor{pink4}{HTML}{C2726E}
\definecolor{pinkintro}{HTML}{F3D6D6}

\definecolor{darkblue}{rgb}{0, 0, 0.5}
\hypersetup{colorlinks=true, citecolor=darkblue, linkcolor=darkblue, urlcolor=darkblue}

\def\ours{\text{STAT}}
\def\sel{\text{STAT-Sel}}
\def\syn{\text{STAT-Syn}}
\def\consel{\text{STAT-ConSel}}
\def\consyn{\text{STAT-ConSyn}}

\def\naive{MATH-Train}
\def\augmented{MATH-Augment}
\def\hard{MATH-Hard}
\def\embedsel{Embed-Sel}
\def\embedsyn{Embed-Syn}
\def\difficult{MATH$^{D}$}
\def\phard{MATH-perturb-hard}

\def\profile{\texttt{Missing\text{-}Skill\text{-}Profile}}
\def\skillmap{\texttt{Skill\text{-}Map}}

\def\test{$\mathcal{Q}$}
\def\train{$\mathcal{P}$}
\def\ptarget{$\mathcal{P}_{targeted}$}
\def\skills{$\mathcal{S}$}
\def\qval{$\mathcal{Q}^{val}$}
\def\qtest{$\mathcal{Q}^{test}$}
\def\diffval{$\mathcal{Q}_{\text{difficult}}^{val}$}

\def\math{MATH}
\def\gsm{GSM8K}

\def\qwM{Qwen2.5-3B}
\def\qwS{Qwen2.5-1.5B-Instruct}
\def\llamaM{Llama-3.2-3B-Instruct}
\def\llamaS{Llama-3.2-1B-Instruct}

\def\gpt{GPT-4o-mini}

\newcommand{\cmark}{\large \textcolor{sage5}{\ding{51}}} 
\newcommand{\cross}{\large \textcolor{pink3}{\ding{55}}}   

\newcolumntype{M}[1]{>{\arraybackslash}m{#1}} 
\newcolumntype{P}[1]{>{\raggedright\arraybackslash}p{#1}} 

\newtcolorbox{casestudy}[3][]{
    colback=#2,
    colbacktitle=#3,
    colframe=#3,
    coltitle=white,
    fontupper=\small,
    fonttitle=\bfseries,   
    boxsep=2pt,
    left=0pt,
    right=0pt,
    top=0pt,
    bottom=0pt,
    boxrule=0.8pt,
    enhanced,
    breakable,
    center title,          
    #1,                    
}

\newcommand{\prompttext}[1]{\textcolor{black}{#1}}

%% file: main_sections/abstract.tex
\begin{abstract}
Language models often show little to no improvement (i.e., ``saturation'') when trained via vanilla supervised fine-tuning (SFT) on data similar to what they saw in their training set (e.g., MATH).
We introduce a new fine-tuning strategy, \ours{}, to train such a student model by using the metacognition ability of a stronger large language model (LLM) as the teacher.
 The teacher uses the task dataset to create a list of skills needed for the task, and then labels each data point with its required skills \citep{didolkar2024metacognitive}. By monitoring the student’s answers, the teacher creates a \profile{} for the student, tracking how often they failed to apply each skill in their responses. We use this idea to build a modified training set in one of two ways. In {\em \sel{}}, the teacher uses an existing set of training examples but adaptively reweights them according to the \profile{}.
In {\em \syn{}}, the teacher synthesizes additional examples involving missing skills. Across extensive experiments on Llama and Qwen models, our methods yield improvements of up to $7.5\%$ on \math{}, whereas SFT provides only limited gains. Furthermore, \ours{} enhances performance on out-of-distribution benchmarks (e.g., AIME24/25, AMC23, etc.) by an average of $4.6\%$.  Crucially, we find that \ours{} is complementary to RL via GRPO~\citep{shao2024deepseekmath}: after the model is improved using \ours{} to address skill gaps, GRPO continues to add further gains. We conclude that skill-targeted adaptive training should broadly improve current training pipelines. \footnote{\faGithub~\href{https://github.com/princeton-pli/STAT}{https://github.com/princeton-pli/STAT}.}
\end{abstract}

\begin{table*}[htbp]
    \centering
    \fontsize{8.5pt}{10pt}\selectfont
    \setlength{\tabcolsep}{3.8pt}
    \begin{tabular}{l@{\hspace{-1pt}}cc@{\hspace{2pt}}c@{\hspace{2pt}}c@{\hspace{2pt}}cccccc}
        \toprule
         \multirow{2}{*}{Models} 
         & \multirow{2}{*}{\textbf{MATH}} 
         & \multirow{1.85}{*}{\textbf{MATH$^{\mathbf{D}}$}} 
         & \multirow{1.88}{*}{\textbf{MATH$^{\mathbf{2}}$}}  
         & \multirow{2}{*}{\textbf{GSM8K}} 
         & \multirow{2}{*}{\textbf{AMC23}} 
         & \multicolumn{2}{c}{\textbf{MATH-perturb}} 
         & \multicolumn{2}{c}{\textbf{AIME}}  
         & \multirow{2}{*}{\textbf{Avg.}} \\
  \cmidrule(lr){7-8} \cmidrule(lr){9-10}
    & & & & & & \textbf{simple}& \textbf{hard} & \textbf{2024}  &\textbf{2025}\\
        \midrule
    \llamaM{} & 44.0 & 18.2 & 21.9 & 73.0 & 21.7 & 33.7 & 12.2 & 33.3 & 16.7 & 30.5\\
    {\quad +SFT}      & 44.8 & 22.9 & 21.0 & 75.1 & 20.8 & 33.0 & 12.2 & 30.0 & 20.0 & 31.1\\
    {\quad +GRPO}   & 45.4 & 24.4 & 23.3 & 77.4 & 25.8 & 38.4 & 11.8 & 33.3 &  6.7 & 31.8\\
    
    \textbf{\sel} & \textbf{51.5} & 26.6 & 25.7 & \textbf{80.2} & \textbf{24.7} & \textbf{39.8} & 13.3 & \textbf{43.3} & 23.3 & {\cellcolor{sage2}\textbf{36.5}}\\
    \textbf{\syn} & 50.2 & \textbf{31.7} & \textbf{26.2} & 79.2 & 23.9 & 39.1 & \textbf{14.7} & 40.0 & \textbf{30.0} & {\cellcolor{pinkintro}\textbf{37.2}}\\
    
    \bottomrule
\end{tabular}
\captionsetup{font=small}
    \caption{\ours{} significantly enhances the performance of \llamaM{} on various math benchmarks by targeting its missing skills in solving \math{}. 
    See \Cref{tab:main-results} for full evaluation results.}
    \vspace{-5pt}
    \label{tab:head-results}
\end{table*}

%% file: main_sections/introduction.tex
\section{Introduction}

Language models have demonstrated remarkable success at acquiring knowledge from large-scale natural text corpora through the next-token prediction objective \citep{shannon1951prediction}. Subsequent supervised fine-tuning on curated data using the same objective then leads to strong performance on domain-specific tasks such as mathematics. However, this process is often inefficient and data hungry \citep{kaplan2020scaling,muennighoff2023scaling,zhang2024scaling,villalobos2024position}, with models quickly reaching a {\em saturation point} for a fixed dataset whereby further training does not help performance. Several works have suggested that this saturation happens because the loss is an average over data points, causing the training signal to diminish as the model becomes adept at most of the training examples~\citep{chen2023skill,xie2023doremi,lin2024rho,tong2024dartmath,jiang2024adaptive,xue2025dast,zhang2025d3}. In addition, there is a mismatch between the ``average'' next-token prediction loss used during training and the auto-regressive generation process used to evaluate performance. As a result, the average loss may fail to capture the specific generation errors that remain in a saturated model~\citep{arora2022exposure,fang2024wrong}.

\looseness-1To tackle this saturation, prior works have shown that adapting the training data distribution can boost performance on inference-time tasks. The key idea is to focus the next-token prediction loss on an adapted set of examples targeted towards good generation \citep{xia2024less,yu2024mates,lin2024rho}.  
This is primarily done by using embeddings or gradient-based estimates to pick training examples most relevant to reducing loss on a reference validation set. 
While these methods show benefits, they remain anchored to validation-set loss, which is only a coarse proxy for a model’s actual generation errors. In fact, our experiments reveal that embedding-based methods, which adapt training data by measuring similarity to validation questions the model fails on, can be ineffective (\Cref{sec:results}) on saturated models that have undergone extensive post-training, e.g., Llama-instruct models. 

We propose to address the saturation problem by drawing inspiration from pedagogical practices rooted in cognitive science, which customize training to specially target the student's skill-deficiencies~\citep{bandura1977social,hattie2007power}.

\begin{quote}
{\em How can we effectively use today's strong teacher models to design better training strategies to help small models overcome their saturation? }
\end{quote}

\begin{figure}[t]
    \centering
    \includegraphics[width=\linewidth]{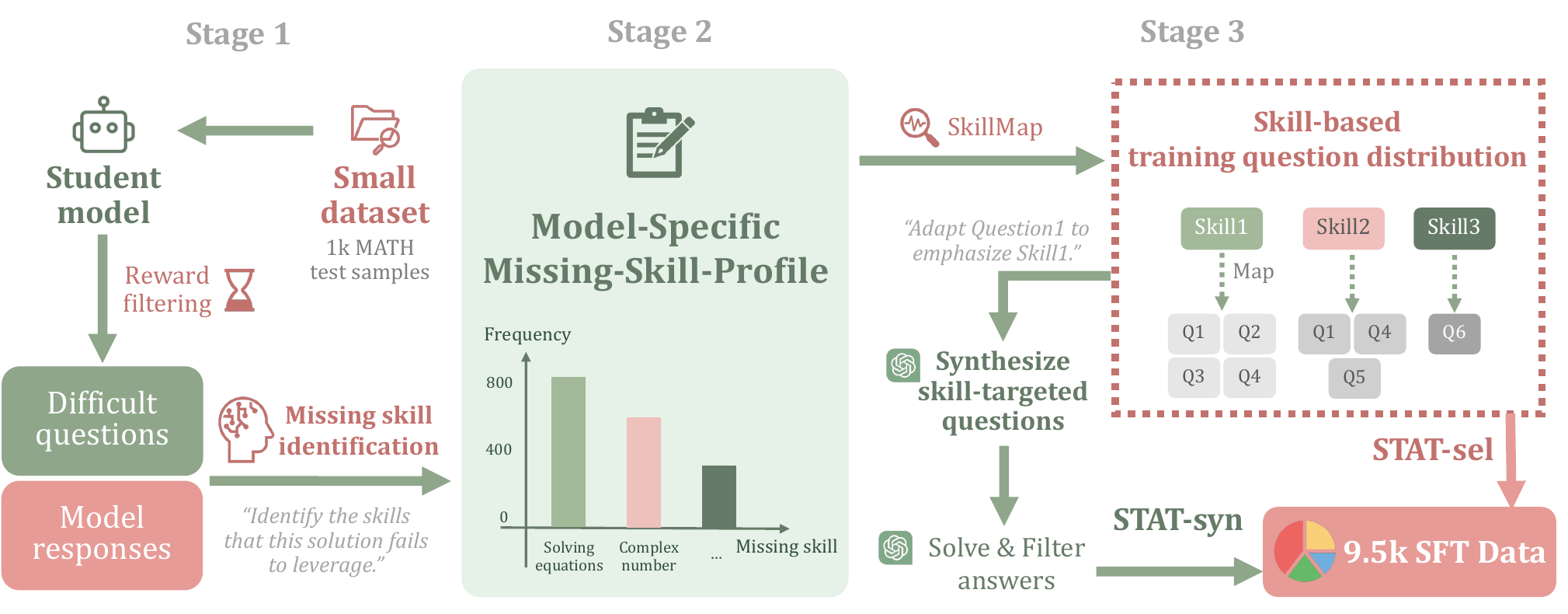}
    \vspace{-8pt}
    \caption{\ours{} is a three-stage skill-based data selection/generation method for supervised fine-tuning (SFT). \textbf{Stage 1:} Identify difficult questions for each model using reward filtering on model responses. \textbf{Stage 2:} Use frontier LLMs to analyze the model responses and build a model-specific \profile{}. \textbf{Stage 3:} Use a pre-constructed \skillmap{} to map the missing skill distribution to a training question distribution, which constitutes the \sel{} data. \syn{} synthesizes new questions using frontier LLMs targeted to the missing skills.} 
    \vspace{-10pt}
    \label{fig:figure1}
\end{figure}

We turn to a growing line of research in LLM meta-cognition \citep{didolkar2024metacognitive,kaur2024instruct}, which leverages the predictive abilities of frontier LLMs to reason about the high-level skills required to solve a given task, as well as the skills actually being used in the student's answer. Thus, in principle, frontier LLMs can act  as 
\underline{the teacher} who guides the training process of \underline{the student} model, actively monitoring the student's competence on individual skills and adjusting their training examples.

\textbf{Informal description of data design:} 
\looseness-1
Our pipeline starts with a list of relevant skills for the problem created via teacher metacognition~\citep{didolkar2024metacognitive}, and adds three stages. First, we use the teacher to evaluate the student model on a small validation set of questions and use a reward model to identify the questions that are difficult for the student. Second, we create a \profile{} by using the teacher to check the missing skills in the model responses. Our first method \textbf{\sel{}} simply up-weights training examples using the  \profile{}; in effect, this guides the student to focus on their deficiencies.  Our second (more expensive) method \textbf{\syn{}} uses the teacher to generate synthetic training data using in-context examples from the validation set associated with a list of deficient skills in  \profile{}. 

\paragraph{Key findings:} Applying \sel{} and \syn{} teaching on Llama and Qwen models with the popular MATH \citep{hendrycks2021measuring} data shows the following:
\begin{enumerate} 
    \item \textbf{Substantial in-distribution gains:} \ours{} achieves improvement on \math{} by up to $7.5\%$, whereas naive fine-tuning yields negligible gains. Previous embedding-based data selection strategies that adapt to the student's validation errors are found ineffective (\Cref{sec:results}).
    
    \item \textbf{Strong out-of-distribution (OOD) generalization:} Improvements in difficult and OOD benchmarks such as AIME24/25 and AMC23 highlight the general utility of skill-targeted training (\Cref{sec:results}).

    \item \textbf{Adaptivity to evolving tasks:} Extending the previous observation, we show that \sel{} and \syn{} can be continually adapted to new, harder evaluation settings, i.e., new validation set, while still leveraging the same training set (\Cref{sec:continual_train}).
    \item \textbf{Supplementary benefits over reinforcement-learning (RL):} We show that  STAT followed by RL improves upon RL-only training, such as GRPO \citep{shah2024ai} (\Cref{sec:results}). This suggests that STAT is likely to prove relevant to most training pipelines today.
\end{enumerate}

We conduct extensive ablations to pinpoint the success of our proposed methods. A fine-grained skill-level analysis reveals that even when models have been extensively trained on \math{}, they struggle on basic operation skills such as basic algebra (\Cref{sec:discussion}). By explicitly targeting these basic skills, our methods reduce such errors and improve generation performance, including on out-of-distribution tasks. In contrast, alternative approaches such as embedding-based methods often underperform, as they do not explicitly address these skill gaps (\Cref{fig:skill-profile-contrast}). Thus, our findings demonstrate the robustness and broad applicability of skill-aware targeted training.

%% file: main_sections/preliminary_design.tex
\section{\ours{}: Adapting training to model's missing skills}\label{sec:design}

Let $\mathcal{Q}$ be the set of test-set questions, out of which we use a subset \qval{} as validation data and \qtest{}$=\mathcal{Q}\setminus$\qval{} as evaluation data. We also have access to a set of training questions $\mathcal{P}$,  which has been utilized to train the student model during its pre-training or post-training phase, and naively fine-tuning the model on $\mathcal{P}$ offers little to no improvement. In our experiments, we use the test and training dataset from \math{}, denoted as $\mathcal{Q}$ and $\mathcal{P}$ respectively. 
We aim to build a targeted training dataset $\mathcal{P}_{targeted}$ to train the model further.

Our work builds on using metacognitive abilities of frontier models from \cite{didolkar2024metacognitive}, which we describe here. While hard to define precisely, a skill is informally defined as a basic computation necessary to solve a task at hand. For example, necessary skills to solve arithmetic tasks could be addition, subtraction, multiplication and division. 
We will use $\mathcal{S}$, a set of skills that are necessary to solve questions in $\mathcal{Q}$ and $\mathcal{P}$. These skills are enlisted from a large model like GPT-4o  using an appropriate prompting strategy \citep{didolkar2024metacognitive,kaur2024instruct}. Then, we create \skillmap{}$: \mathcal{S} \to \mathcal{P}$ to be a map from a skill to the set of training questions that require applying the skill, which we will also get by prompting the same LLM \citep{achiam2023gpt}. We use the skill set $\mathcal{S}$ and the \skillmap{} from \cite{didolkar2024metacognitive}.

To develop \sel{} and \syn{}, we first identify difficult questions for the student model on a validation set by analyzing its own responses. For these questions, we then use the teacher model to infer the skills that are missing from the student’s reasoning. A skill-targeted training set is constructed by emphasizing examples corresponding to these missing skills, either via up-weighting samples or synthesizing new ones. Unless otherwise specified, all of our experiments use \gpt{} as the teacher model.

\subsection{Stage 1: Detection of {\em difficult} questions via reward filtering}
\label{sec:stage1}

In this stage, we will label a question $q \in \mathcal{Q}$  as {\em difficult} or not for the student model. We could simply define {\em difficult} questions as the set of questions that the model gets wrong after evaluation. However, this requires access to the ground truth labels. Instead, to make our technique more broadly applicable, we use a reward model to classify the responses of the student model. The reward model need not be a perfect reward model; we give more ablations in \Cref{app:prm-ablations}. Given a question $q$, we use a reward model to score the response of the student model.

\textbf{Reward filtering.} As we primarily focus on math datasets, we assume that the model's response is composed of $t$ steps for a question $q$ and contains the answer in its final step. We will use the reward model to output reward scores for each step. For simplicity, we will refer to the scores of the reward model as $\{ r_{q, 1}, \cdots, r_{q, t} \}$. Then, we use thresholds $\tau_1, \tau_2$ to filter out {\em difficult} questions for the student model. We will refer to the threshold filtering function as $R: \mathcal{Q} \to \{0, 1\}$.

\begin{equation}
R (q) = 
\begin{cases}
0, & \begin{aligned}[t]
        &\text{(if) }  r_{q, t} \leq \tau_1 
        & \quad \text{(final step has low reward)} \\
        & \text{(or) } \frac{1}{t} \sum\limits_{i=1}^{t} r_{q, i} \leq \tau_1 
        &\quad \text{(average low reward across all steps)} \\
        & \text{(or) } \exists i < t \text{ s.t. } r_{q, i} \leq \tau_2 
        & \quad \text{(low reward at any step)} 
    \end{aligned} \\
1, & \text{otherwise},
\end{cases} \label{eq:reward}
\end{equation}

\textbf{Identifying difficult questions.}
We define $\mathcal{Q}_{\text{difficult}}$ as a model-specific subset of the \math{} dataset, consisting of questions with low-reward model responses $R$. 
To avoid training directly on the test data, we use two splits of $\mathcal{Q}_{\text{difficult}}$ as validation and test sets: 

\quad $\bullet$ $\mathbf{\mathcal{Q}_{difficult}^{val}}$: Difficult questions in the validation set, given by $\mathcal{Q}_{\text{difficult}} \cap \mathbf{\mathcal{Q}^{val}}$, are used to label missing skills in Stage 2.

\quad $\bullet$ $\mathbf{\mathcal{Q}_{difficult}^{test}}$: 
Difficult questions in the test set, given by $\mathcal{Q}_{\text{difficult}} \cap \mathbf{\mathcal{Q}^{test}}$, are used for \difficult{} evaluation  in \Cref{tab:main-results}.

\subsection{Stage 2: Constructing model-specific \profile{}} \label{sec:stage2}

For each {\em difficult} question $q$ in \diffval{}, we use a frontier LLM (\gpt{}) to predict the set of skills in $\mathcal{S}$ that are missing in the model's responses. We call this map \profile{}$: \text{\diffval{}} \to \mathcal{S}$. This map will be used to build \sel{} and \syn{}. See \Cref{sec:discussion} for examples and an extensive analysis of \profile{} across models, and \Cref{app:prompts} for prompts.

\subsection{Stage 3: Selecting or synthesizing skill-based training data}
\label{sec:stage3}
In this stage, we construct our skill-targeted training dataset, $\mathcal{P}_{targeted}$.

\textbf{\sel{}.} We create this set by directly sampling questions from the training dataset $\mathcal{P}$ according to the skills listed in the \profile{}. Specifically, for each question $q \in \text{\diffval{}}$, we examine \profile{}(q) and, for every skill it contains, sample multiple questions from $\mathcal{P}$ that are linked to the same skill via the \skillmap{}. Consequently, the frequency with which a skill contributes to the selection process is proportional to the number of questions associated with that skill in the \profile{}.

\textbf{\syn{}.} We generate new synthetic questions using the teacher model. For each question $q \in \text{\diffval{}}$, we examine \profile{}(q). For each skill it contains, we randomly sample $3$ questions from $\mathcal{P}$ that are linked to the same skill via the \skillmap{}, and ask the teacher model to create new questions and responses by referring to the sampled questions. We keep only those questions where the teacher model is consistent across at least $2$ of its responses, and keep only those question-answer pairs in our training set. 
Detailed procedures are given in \Cref{app:data-creation-details}.

%% file: main_sections/experiments.tex
\section{Experiments}

\subsection{Experimental Setup} \label{sec:expt-set-up}
\looseness-1\textbf{Datasets.}
All training data for \ours{} and the baselines are either selected or synthesized from the \math{} dataset (7.5k train / 5k test) \citep{hendrycks2021measuring}. In addition to the original solutions provided in the dataset, we also collect three alternative versions of each answer by prompting the teacher model to re-write them three times. We further report performance of \ours{} and each baseline after continuing training with GRPO \citep{shao2024deepseekmath} on the same MATH training set. We randomly split the MATH test set into 1k validation and 4k test subsets, with both \math{} and \difficult{} evaluations drawn from the 4k test split. See \Cref{sec:stage1} for design details.
We also evaluate our method on extensive OOD benchmarks including \gsm{} \citep{cobbe2021training}, MATH$^2$ \citep{shah2024ai}, MATH-perturb \citep{huang2025math}, AMC23 \citep{aimo-validation-amc2025}, and AIME2024/2025 \citep{huggingfaceh4_aime_2024,huggingfaceh4_aime_2025}. 


\looseness-1\textbf{Model \& Training Configuration.} 
We focus on smaller models as a testbed, as their performance remains noticeably weaker on \math{}. We employ \gpt{} as the teacher model, and apply \ours{} on student models \llamaM{}, \llamaS{} \citep{MetaAI2024}, and \qwM{} \citep{qwen2025qwen25technicalreport}, and evaluate under 0-shot settings.
We fine-tune models for 3 epochs, with learning rate chosen separately for each method based on accuracy on \math{}. We provide detailed hyperparameters in \Cref{app:model-training-config}, ablations on threshold sensitivity in \Cref{app:prm-ablations}, and a discussion of teacher model variants in \Cref{app:teacher-analysis}.

\newcommand{\vcentercell}[2]{\raisebox{\dimexpr #1\height}{\makecell[l]{#2}}}
\begin{table}[t]
\centering
\small
\setlength{\tabcolsep}{2.2pt}
\fontsize{8.5pt}{10pt}\selectfont
\begin{tabular}{l@{\hspace{-4pt}}cccp{7.5cm}}
\toprule
\multirow{2}{*}{\textbf{Method}} &
\textbf{\# Unique} &
\textbf{\# QA} &
\textbf{Synthetic} &
\multicolumn{1}{c}{\parbox[c]{7.5cm}{\centering \textbf{Training Data}}} \\
& \textbf{Questions} & \textbf{Pairs} & \textbf{Data} & \multicolumn{1}{c}{\parbox[c]{7.5cm}{\centering \textbf{Description}}}\\
\midrule
\vcentercell{0.1}{\naive{} \\ \citep{hendrycks2021measuring}}
  & \vcentercell{0.4}{7.5k} & \vcentercell{0.4}{7.5k} & \vcentercell{0.4}{\cross{}} & \vcentercell{0.2}{\math{} original training set.} \\

\hline
\vcentercell{-0.5}{\augmented{} \\ \citep{mathplus2024}} 
  & \vcentercell{-0.8}{7.5k} & \vcentercell{-0.8}{9.5k} & \vcentercell{-0.8}{\cross{}} & Augmented \math{} training set with multiple teacher-rewritten solutions per question. \\

\hline
\vcentercell{-0.6}{\makecell[l]{\hard{} \\ \citep{sun2024easytohardgeneralizationscalablealignment}}}
  & \vcentercell{-1}{3k} & \vcentercell{-1}{9.5k} & \vcentercell{-1}{\cross{}} & \vcentercell{-1}{Subset of \augmented{} with Level 4–5 \math{} questions.} \\

\hline
\vcentercell{-0.6}{\embedsel{} \\ \citep{li2025massmathematicaldataselection}} 
  & \vcentercell{-1.1}{4k} & \vcentercell{-1.1}{9.5k} & \vcentercell{-1.1}{\cross{}} & Reweighted set of \augmented{} via upweighting training questions similar to the {\em difficult} questions in embedding space. \\

\hline
\vcentercell{-0.6}{\embedsyn{} \\ \citep{jung2025prismatic}} 
  & \vcentercell{-1}{4k} & \vcentercell{-1}{9.5k} & \vcentercell{-1}{\cmark{}} & 
  Synthetic \math{}-level QAs generated by the teacher, using training examples from \embedsel{} as references.
  \\

\hline
\vcentercell{-0.9}{\textbf{\sel{}} \\ (Ours)} 
  & \vcentercell{-1.3}{4k} & \vcentercell{-1.3}{9.5k} & \vcentercell{-1.3}{\cross{}} & Reweighted set of \augmented{} via upweighting training questions related to model's missing skills in solving the {\em difficult} questions.
  \\

\hline
\vcentercell{-1}{\textbf{\syn{}} \\ (Ours)} 
  & \vcentercell{-1.4}{4k} & \vcentercell{-1.4}{9.5k} & \vcentercell{-1.4}{\cmark{}} & 
   Synthetic \math{}-level QAs generated by the teacher, with training examples from \sel{} and their associated skills as references. \\
\bottomrule
\end{tabular}
\caption{Comparison of training data construction methods. We attach details of data construction procedure in \Cref{app:data-creation}.}
\vspace{-10pt}
\label{tab:baseline_details}
\end{table}

\begin{table*}[t]
    \centering
    \small
    \fontsize{8.5pt}{10.5pt}\selectfont
    \setlength{\tabcolsep}{4.7pt}
    \begin{tabular}{lcc@{\hspace{2.3pt}}c@{\hspace{2.3pt}}c@{\hspace{3pt}}cccccc}
        \toprule
         \multirow{2}{*}{Methods} 
         & \multirow{2}{*}{\textbf{MATH}} 
         & \multirow{1.9}{*}{\textbf{MATH$^{\mathbf{D}}$}} 
         & \multirow{1.9}{*}{\textbf{MATH$^{\mathbf{2}}$}}  
         & \multirow{2}{*}{\textbf{GSM8K}} 
         & \multirow{2}{*}{\textbf{AMC23}} 
         & \multicolumn{2}{c}{\textbf{MATH-perturb}} 
         & \multicolumn{2}{c}{\textbf{AIME}}
         & \multirow{2}{*}{\textbf{Avg.}} 
         \\
         \cmidrule(lr){7-8} \cmidrule(lr){9-10}
        & & & & & & \textbf{simple}& \textbf{hard} & \textbf{2024}  &\textbf{2025}\\
        \midrule
        \rowcolor{sage1}
        \multicolumn{11}{c}{\textit{\llamaM{} $+$ SFT}}\\
    Base Model & 44.0 & 18.2 & 21.9 & 73.0 & 21.7 & 33.7 & 12.2 & 33.3 & 16.7& 30.5\\
    
        \naive{}  & 44.8 & 22.9 & 21.0 & 75.1 & 20.8 & 33.0 & 12.2 & 30.0 & 20.0& 31.1\\
        \augmented{}  & 45.2 & 23.9 & 23.8 & 77.8 & 23.8 & 35.1 & 12.5 & 30.0 & 13.3& 31.7\\
        \hard{}  & 45.6 & 24.9 & 23.3 & 78.2 & 21.6 & 38.0 & 11.8 & 30.0 & \underline{26.7}& 33.3\\
        \embedsel{} & 46.0 & 26.5 & 20.5 & 76.6 & 21.6 & 36.2 & \textbf{14.7} & 36.7 & 16.7& 32.8\\
        
        \embedsyn{}  & 48.8 & \underline{27.3} & 19.5 & 78.4 & 22.7 & 36.9 & \underline{13.3} & 26.7 & 23.3 & 33.0\\
        
        \textbf{\sel} & \textbf{51.5} & 26.6 & \underline{25.7} & \textbf{80.2} & \textbf{24.7} & \textbf{39.8} & \underline{13.3} & \textbf{43.3} & 23.3& {\cellcolor{sage2}\underline{36.5}}\\

        \textbf{\syn}  & \underline{50.2} & \textbf{31.7} & \textbf{26.2} & \underline{79.2} & \underline{23.9} & \underline{39.1} & \textbf{14.7} & \underline{40.0} & \textbf{30.0}& {\cellcolor{pinkintro}\textbf{37.2}}\\
        
        \rowcolor{lightgray}
        \multicolumn{11}{c}{\textit{$+$ GRPO}}\\
    Base Model & {45.4} & {24.4} & {23.3} & {77.4} & {25.8} & {38.4} & {11.8} & {33.3} & {3.3}& 31.8\\
        
        {\naive{}} & {46.4} & {28.4} & {28.6} & {80.7} & {29.7} & {37.6} & {12.5} & {36.7} & {10.0}& 34.5\\
        
        {\augmented{}} & {47.4} & {31.6} & {28.6} & {81.4} & {30.6} & {37.6} & {14.0} & {36.7} & {\textbf{33.3}}& 37.9\\
        
        {\hard{}} & {49.4} & {33.2} & {28.6} & {80.3} & {31.3} & {39.1} & {15.4} & {\underline{43.3}} & {13.3}& 37.1\\
        
        {\embedsel{}} & {50.4} & {37.5} & {23.8} & {80.5} & {32.0} & {38.0} & {\underline{16.8}}  & {36.7} & {20.0}& 38.8\\
        {\embedsyn{}} & {49.7} & {\underline{37.8}} & {19.5} & {80.6} & {33.9} & {39.1} & {\underline{16.8}} & {36.7} & {23.3}& 38.6\\

        {\textbf{\sel{}}} & {\textbf{52.2}} & {35.0} & {\textbf{32.4}} & {\underline{81.8}} & {\textbf{34.2}} & {\underline{42.7}} & {\textbf{17.6}} & {\underline{43.3}} & {\underline{26.7}}& {\cellcolor{sage2}\underline{40.7}}\\
        {\textbf{\syn{}}} & {\underline{51.0}} & {\textbf{39.1}} & {\underline{29.0}} & {\textbf{82.0}} & {\underline{31.9}} & {\textbf{43.0}} & {15.8} & {\textbf{46.7}} & {\textbf{33.3}}& {\cellcolor{pinkintro}\textbf{41.3}}\\
        \midrule
        \rowcolor{sage1}

\rowcolor{sage1}
\multicolumn{11}{c}{\textit{Qwen2.5-3B $+$ SFT}}\\

Base model   & 55.8 & 45.3 & 34.8 & 80.9 & 26.4& 43.7 & \underline{24.0}  & 23.3 & 20.0 & 39.4 \\
\naive{}     & 50.0 & 44.2 & 32.9 & 80.1  & 33.6& 42.3 & 23.3 & 26.7 & \underline{26.7} & 40.0 \\
\augmented{} & 56.6 & 45.6 & 37.1 & 80.4& {33.0} & 40.9 & 21.9  & 16.7 & \underline{26.7} & 39.9 \\
\hard{}      & 56.7 & 45.6 & 31.4 & 79.8 & {33.6}& 43.7 & 23.7  & 30.0 & 16.7 & 40.1 \\
\embedsel{}  & 57.5 & 46.4 & 34.3 & 80.4 & {33.6}& 43.7 & 21.9  & 30.0 & \underline{26.7} & 41.6 \\
\embedsyn{}  & 56.4 & 47.4 & 34.3 & 80.4 & \underline{35.2}& 43.7 & \underline{24.0}  & 26.7 & \underline{26.7} & 41.6 \\
\textbf{\sel} & \underline{58.4} & \underline{47.6} & \underline{39.5} & \textbf{82.3} & \textbf{35.5}& \textbf{45.9} & \underline{24.0} & \underline{33.3}  & \textbf{30.0} & {\cellcolor{sage2}\underline{44.1}} \\
\textbf{\syn} & \textbf{59.4} & \textbf{49.2} & \textbf{40.5} & \underline{81.3} & 34.4 & \underline{44.8} & \textbf{25.1} & \textbf{36.7} & \textbf{30.0} & {\cellcolor{pinkintro}\textbf{44.6}} \\

\rowcolor{lightgray}
\multicolumn{11}{c}{\textit{$+$ GRPO}}\\
Base model & 61.6 & 49.8 & 41.0 & \underline{85.1} & 37.7& 49.8 & 25.8  & 33.3 & 30.0 & 46.0 \\
\naive{} & 61.6 & 51.1 & 34.8 & 84.8 & 36.9 & \textbf{51.6} & 26.5 & 33.3 & 30.0 & 45.6 \\
\augmented{} & 61.0 & 48.2 & 40.5 & 84.0  & 36.3& 48.7 & 26.2 & \underline{36.7} & 26.7 & 45.4 \\
\hard{} & 59.0 & 51.1 & 35.7 & 84.2& 37.7 & 49.8 & 26.5  & 33.3 & 23.3 & 44.5 \\
\embedsel{} & 59.7 & 48.9 & 41.0 & 84.3 & 38.4& 46.6 & 25.8  & 26.7 & \textbf{36.7} & 45.3 \\
\embedsyn{} & 61.4 & \underline{52.3} & 40.0 & 83.7 & \underline{38.8}& 47.7 & \underline{28.0}  & 26.7 & 30.0 & 45.4 \\
\textbf{\sel} & \textbf{62.8} & 52.1 & \textbf{44.8} & 84.8 & \underline{38.8}& 48.7 & \textbf{30.1}  & \underline{36.7} & \underline{33.3} & {\cellcolor{sage2}\underline{48.0}} \\
\textbf{\syn} & \underline{61.8} & \textbf{52.4} & \underline{41.9} & \textbf{85.6} & \textbf{39.2}& \underline{50.9} & 26.9  & \textbf{40.0} & \textbf{36.7} & {\cellcolor{pinkintro}\textbf{48.4}} \\
    \bottomrule
\end{tabular}

    \captionsetup{font=small}
    \caption{Improvements on various math benchmarks from applying STAT. 
    Results under ``+SFT" show the performance of SFT models trained with each method, while ``+GRPO" shows the performance after applying GRPO on top of the corresponding SFT models. Our methods, \sel{} and \syn{}, achieve an average gain of up to $6.7\%$ over the base model, with strong OOD performances (AMC23 results reported on average@64, AIME on pass@64). Applying GRPO on top of fine-tuning with \ours{} further enhances these improvements by $\sim$$4\%$. Full results are provided for \llamaS{} in \Cref{tab:llamaS-results}, \Cref{app:additional-results}.
    }
    \vspace{-15pt}
    \label{tab:main-results}
\end{table*}

\textbf{Baselines.}
We compare skill-aware training against several baselines.
We begin with \naive{}, where the model simply trains on the original \math{} responses, and \augmented{}, which substitutes the responses with teacher re-written answers. We also compare against \hard{}, restricting training to only Level 4–5 questions. Finally, to test whether skills really matter in \sel{} and \syn{}, we swap them out for an embedding-based approach\footnote{We use Alibaba-NLP/gte-Qwen2-7B-instruct \citep{li2023generaltextembeddingsmultistage}}, selecting training questions by their similarity to difficult validation questions from Stage 1. Please find a summary in \Cref{tab:baseline_details}.
We have attached detailed data creation procedure in \Cref{app:data-creation} and prompts in \Cref{app:prompts}.

\subsection{Evaluation Results}\label{sec:results}

We present results for \llamaM{} and \qwM{} in \Cref{tab:main-results} and for \llamaS{} in \Cref{tab:llamaS-results}, \Cref{app:additional-results}. We refer to each untrained model as ‘Base Model’. 
Our findings can be summarized as follows.

\begin{figure}[t]
    \centering
    \includegraphics[width=\linewidth]{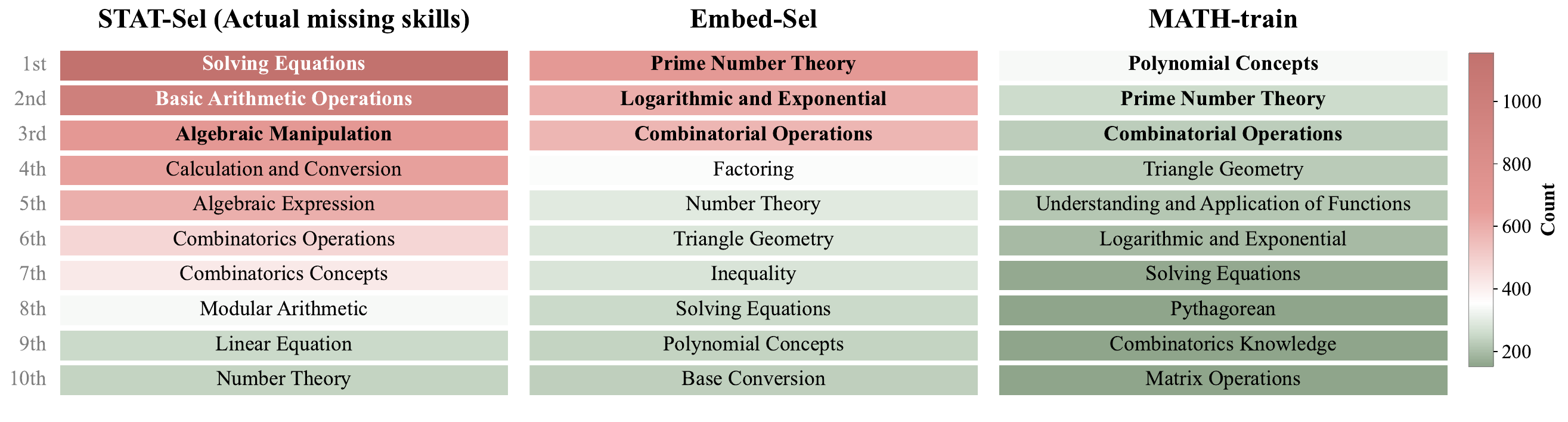}
    \vspace{-15pt}
    \captionsetup{font=small}
    \caption{Comparison among the Top 10 frequent skills present in \sel{}, \embedsel{}, and \naive{} questions selected on \llamaS{}. The skills emphasized in both baselines, \naive{} and \embedsel{}, align poorly with the actual Top 10 missing skills of the model (i.e., skills in \sel{}). Furthermore, the missing skills are not necessarily those most common in the original data distribution, as shown by the skill distribution of \naive{}. } 
    \vspace{-13pt}
    \label{fig:skill-profile-contrast}
\end{figure}
\looseness-1\textbf{Naive SFT provides little to no benefit.} 
Both \naive{} and \augmented{} yield at most a 1--2$\%$ gain over the base model, 
showing that naive SFT offers negligible improvements. It is worth noting that we have systematically tuned hyper-parameters for naive SFT (details attached in \Cref{app:model-training-config}).
In fact, we observe that \qwM{} can even degrade under \naive{}. 
Restricting supervision to only the most difficult \math{} questions (Levels 4--5) 
also fails to produce meaningful gains. 
A natural idea is then to adapt training toward the model’s mistakes by selecting training questions 
semantically close to difficult validation examples. 
Using embedding similarity, \embedsel{} achieves only marginal over \naive{} and \augmented{}. 
Synthetic augmentation via \embedsyn{} provides a small additional boost, but the overall gains remain modest. 




\looseness-1\textbf{Skill-targeted adaptive training shows substantial improvements.} 
\ours{} achieves average gains of up to $6.7\%$ on \llamaM{}, $5.2\%$ on \qwM{}, and $3.4\%$ on \llamaS{}, over the performance of base model. On closer analysis on \difficult{} test set of questions, we show that \syn{} substantially improves the performance of the model on difficult questions, compared to \sel{}, which leads to improved performance overall for \llamaS{} and \qwM{}.



\looseness-1\textbf{Benefits extend beyond \math{}.} 
On out-of-distribution benchmarks, we observe consistent improvements across $7$ datasets, ranging from simpler problems in \gsm{} to challenging competition sets such as AIME. Specifically, \sel{} and \syn{} improve averaged OOD performances by $5.3\%$ and $5.8\%$ respectively, with \syn{} generally excelling on harder tasks such as AIME and \difficult{}.
This demonstrates that targeting skills generalizes extensively beyond the source training set.

\looseness-1\textbf{Compatibility with GRPO.} 
A natural concern is whether our methods can work well with RL-based methods such as GRPO, 
which typically follows SFT \citep{dubey2024llama,guo2025deepseek}. 
For both Llama and Qwen, improvements from SFT on \ours{} have carried over to subsequent GRPO, yielding average gains of up to $9.5\%$ over GRPO on base model.
Surprisingly, on \llamaS{} and \llamaM{}, where GRPO alone does not work well (improving $\leq$$2.4\%$), SFT alone on \ours{} already produced better results than GRPO, and adding GRPO on top further boosts performance by $\sim$$4\%$.

\subsection{Continual learning on challenging benchmarks}\label{sec:continual_train}

As our earlier results show, \ours{} already generalizes strongly to a wide range of OOD tasks while using only \math{} data for training. But in practice, models often face evaluation settings that grow harder over time. A natural question then is: {\em can we continue adapting the model to these tougher benchmarks while still using similar questions as \math{}?}

\begin{wrapfigure}{r}{0.4\textwidth}
  \centering
  \captionsetup{font=footnotesize}
  \includegraphics[width=0.33\textwidth]{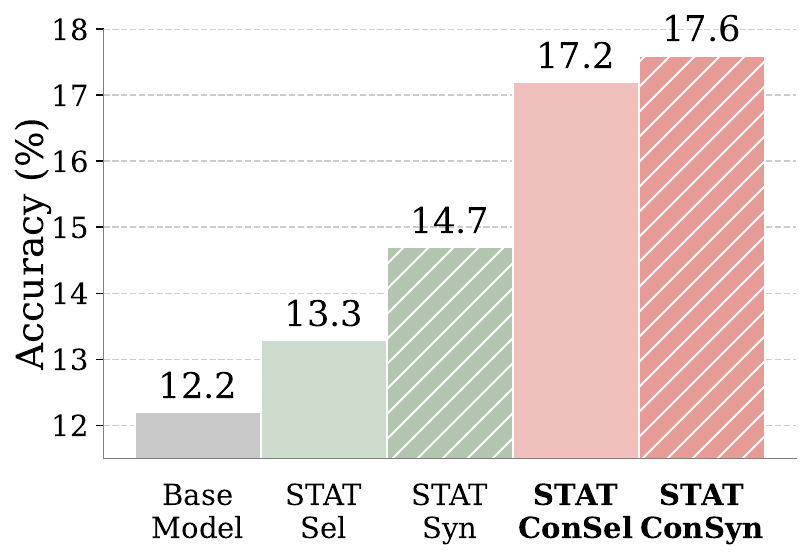}
  \vspace{-5pt}
  \caption{Continual learning results on MATH-perturb-hard. Further fine-tuning STAT models based on their missing skills on unseen data yields a 3--4$\%$ gain (\consel{}/ConSyn).}
  \label{fig:continual-learning}
\vspace{-10pt}
\end{wrapfigure}

For our case study, we consider \phard{}. We report performance for two model variants of \llamaM{}. \textbf{\consel{}} takes a model trained with \sel{}, and trains further with a data creation pipeline identical to \sel{}, but with \profile{}  built on validation questions from \phard{}. \textbf{\consyn{}} builds on \syn{} model with the same idea. In both cases, the evaluation benchmark only gives the skill profile, and the training examples still come from \math{}.

\looseness-1As shown in \Cref{fig:continual-learning}, \sel{} and \syn{} trained models show only 1--2$\%$ improvement on \phard{} over the base model performance, which reflects the difficulty of this benchmark. However, continual trained models show a larger gain of 3--4$\%$. This shows that our framework can be readily adapted to unseen test-time datasets by constructing \profile{} directly on them, while still using \math{} training data. Thus, skill-aware training provides a flexible solution to adapt the models with more challenging evaluations while still relying on existing training datasets.

%% file: main_sections/discussion.tex
\section{Why Skill-Targeted Training Works} \label{sec:discussion}




In this section, we dig into the effectiveness of our proposed skill-aware targeted training. We conduct all the ablations and analyses on \llamaS{} due to limited computational resources.  First, we present the \profile{} across all models. We then show that \ours{} improves the student's performance uniformly across these skills. Finally, we show that the baseline strategies are ineffective because of misalignment in the skill distribution in their proposed training data and the missing skills.

\textbf{Models struggle with basic skills.} First, we closely examine the \profile{} across different models, obtained at the end of Stage 2 (\Cref{sec:stage2}). We present the Top 10 frequently missing skills for each model according to their \profile{} in \Cref{fig:skill-profile-contrast} (Left) and \Cref{fig:skill-profile} (appendix \ref{app:additional-results}). The key observations are:

\quad $\bullet$ \textbf{Algebra-centric skills appear at the top,} e.g., manipulating equations, handling expressions, and solving linear forms. This suggests that even though both Llama and Qwen models achieve high performance on \math{}, they systematically struggle with operation computations. 

\quad $\bullet$ \textbf{Most missing skills are shared across models,} e.g., equation-solving skills and basic arithmetic operations are missing in different model families (Llama and Qwen) and sizes (1B and 3B). However, smaller models show more frequent weaknesses in basic computational skills like arithmetic.

\begin{figure}[t]
    \centering
    \includegraphics[width=\linewidth]{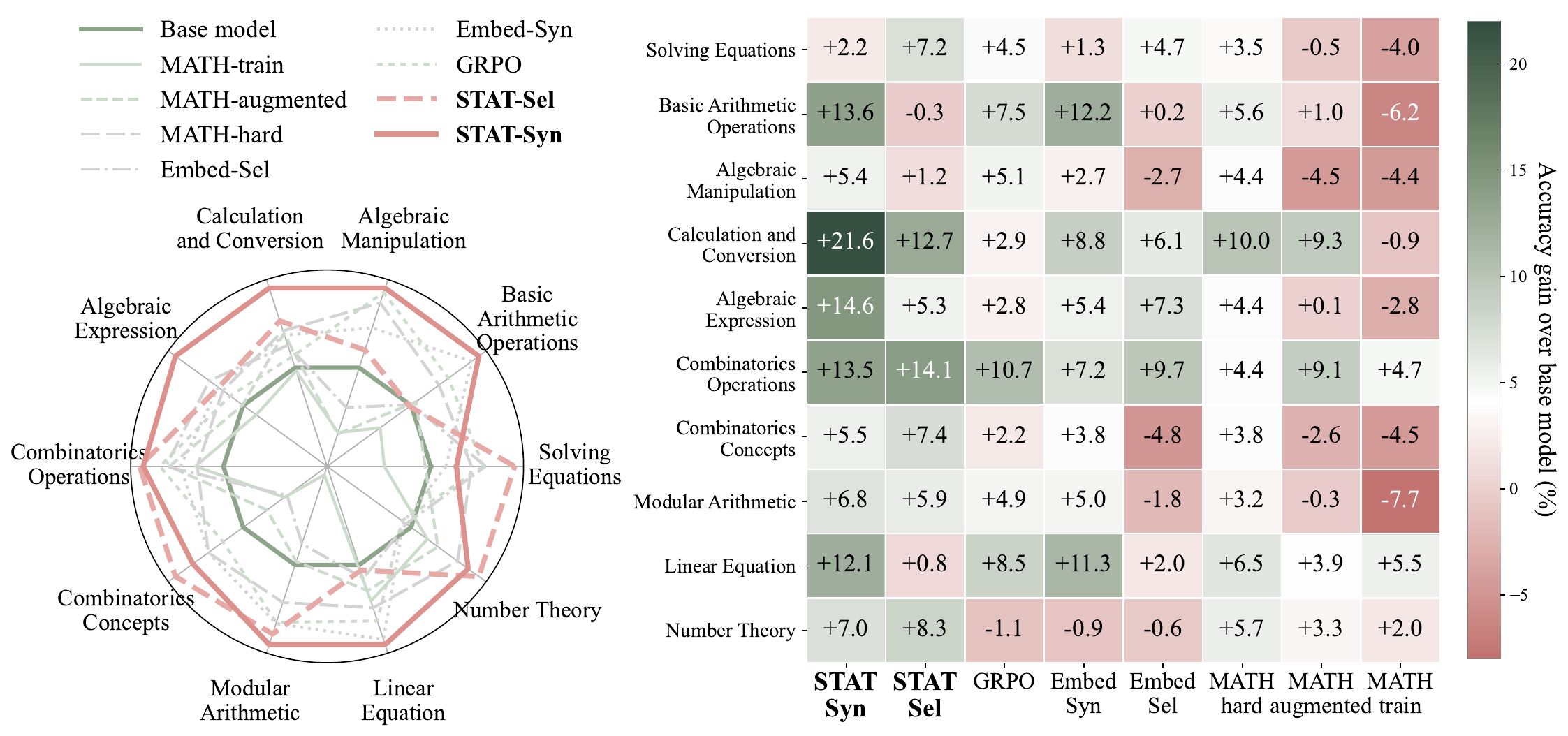}
    \vspace{-8pt}
    \caption{\looseness-1Trained model performances \textbf{(Left)} and performance gain over base model \textbf{(Right)} on Top 10 frequent missing skills, across training strategies on \llamaS{}. Accuracies on the left plot are normalized per skill axis for better visualization. Our approaches \syn{} and \sel{} are most effective in enhancing model performance across nearly all the skills.} 
    \label{fig:skill-improve}
\end{figure}

\textbf{\ours{} effectively addresses models' frequent missing skills:} 
\looseness-1
We take \llamaS{} as a case study to examine how different training strategies impact performance across skills. From its \profile{}, we select the 10 most frequently missing skills and build corresponding evaluation sets, each containing questions annotated via the \skillmap{}. We then measure both absolute performance and performance gains under each method.

As shown in \Cref{fig:skill-improve} (Left), \ours{} consistently outperform all baselines across all 10 skills, whereas baseline models can even fall behind the base model on skills such as Algebraic Manipulation and Modular Arithmetic. \Cref{fig:skill-improve} (Right) provides a quantitative breakdown, showing that \ours{} can deliver over $10\%$ accuracy gains on 5 skills, with the largest improvements on basic skills like Calculation \& Conversion, Algebraic Expression, and Combinatoric Expressions. Notably, \ours{} also brings clear improvements on knowledge-intensive skills such as Number Theory and Combinatorics.

\textbf{Misalignment between baseline training data and missing skills.} To investigate the reason behind the ineffectiveness of our baseline strategies, we adopt a skill-based evaluation by comparing the skill distribution of their training data with the model’s missing skills in the \profile{}. \Cref{fig:skill-profile-contrast} highlights a clear misalignment between the model's actual missing skills (\sel{}) and the baselines: Neither \naive{} nor \embedsel{} addressed the model’s basic algebraic weaknesses, even though \embedsel{} chose data similar to difficult questions by embedding similarity. The skill profile of \naive{} shows a clear gap between missing skills and those skills that occur most common in the training data. This shows that \ours{} effectively targets missing skills, not just the ones that appear most often.
We provide concrete question examples in \Cref{app:question-case-study} to illustrate the distinct differences  between the skills.

\begin{wrapfigure}{r}{0.5\textwidth}
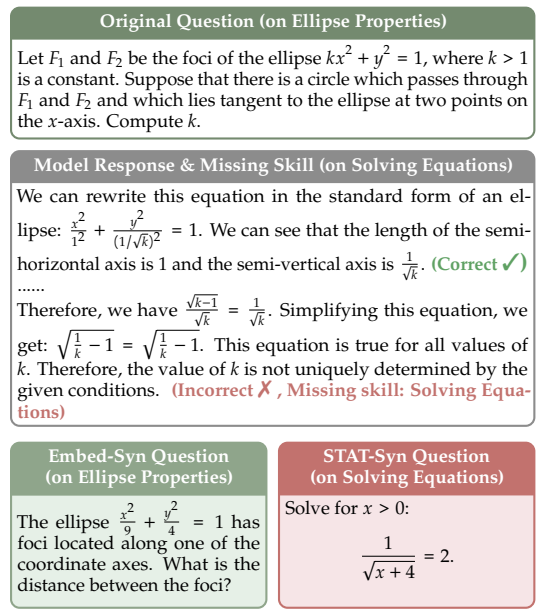

\centering
\scriptsize
\vspace{-3em}
\begin{casestudy}[title={Original Question (on Ellipse Properties)}]{white}{sage5}
\scriptsize
\prompttext{
Let $F_1$ and $F_2$ be the foci of the ellipse $kx^2 + y^2 = 1,$ where $k > 1$ is a constant. Suppose that there is a circle which passes through $F_1$ and $F_2$ and which lies tangent to the ellipse at two points on the $x$-axis. Compute $k.$
}
\end{casestudy}

\begin{casestudy}[title={Model Response \& Missing Skill (on Solving Equations)}]{white}{black!45}
\scriptsize
\prompttext{
We can rewrite this equation in the standard form of an ellipse: $\frac{x^2}{1^2} + \frac{y^2}{(1/\sqrt k)^2} = 1$. We can see that the length of the semi-horizontal axis is $1$ and the semi-vertical axis is $\frac{1}{\sqrt{k}}$. {\textcolor{sagetext}{\textbf{(Correct \footnotesize \ding{51})}}}{\\} 
......{\\}
 Therefore, we have $\frac{\sqrt{k-1}}{\sqrt{k}} = \frac{1}{\sqrt{k}}$. Simplifying this equation, we get:  $\sqrt{\frac{1}{k} - 1} = \sqrt{\frac{1}{k} - 1}$. This equation is true for all values of $k$. Therefore, the value of $k$ is not uniquely determined by the given conditions. { \textcolor{pink4}{\textbf{(Incorrect \footnotesize\ding{55} \scriptsize, Missing skill: Solving Equations)}}}
}
\end{casestudy}
\vspace{-4pt}
{\raggedright
\noindent
\begin{minipage}[t]{0.49\linewidth}
  \begin{casestudy}[equal height group=pairA,title={{\embedsyn{}} Question \\ (on Ellipse Properties)}]{sage1}{sage4}
  \scriptsize
  \prompttext{
  The ellipse $\frac{x^2}{9} + \frac{y^2}{4} = 1$ has foci located along one of the
  coordinate axes. What is the distance between the foci?
  }
  \end{casestudy}
\end{minipage}\hfill
\begin{minipage}[t]{0.49\linewidth}
  \begin{casestudy}[equal height group=pairA,title={{\syn{}} Question \\(on Solving Equations)}]{pink1}{pink4}
  \scriptsize
  \prompttext{
  Solve for $x > 0$:
  \[
  \frac{1}{\sqrt{x+4}} = 2.
  \]}
  \end{casestudy}
\end{minipage}
\par}
\vspace{-5pt}
\captionsetup{font=small}
\caption{Comparison between synthesized questions from \embedsyn{} and \syn{}.}
\label{fig:case-study}
\vspace{-3em}
\end{wrapfigure}




\textbf{Comparing \ours{} to GRPO.}
One of our interesting findings in \Cref{sec:results} was that \ours{} could outperform GRPO training on Llama instruct models. Here, we compare these two approaches from a skill-based perspective. As shown in \Cref{fig:skill-improve} (Right),  although GRPO on \llamaS{} also yields positive gains across nearly all the top skills, the overall effect remains less pronounced compared to \ours{}. A possible reason is that GRPO provides only coarse feedback to the model by contrasting correct and incorrect responses, whereas skill-targeted training pinpoints model weaknesses in a fine-grained way. In light of this, one future direction is to develop a GRPO variant that incorporates skill-based feedback into the reward.

\textbf{Case study on synthetic data.}
To understand why our training samples are skill-targeted, we conduct a case study of the training data.

Here we compare \syn{} with \embedsyn{}, as their data are both created with a specific focus (e.g., embedding-based similarity or missing-skill targeting).

In this example (see \Cref{fig:case-study}), the original question centers on ellipse geometry; the model handles this part well, but showed a gap in the final equation-solving step.
The new question in \embedsyn{}, though highly relevant, captures only the main topic (Ellipse Properties) through embedding similarity.
By contrast, \syn{} leverages the missing-skill information (Solving Equations) and generates a targeted question.

This case study demonstrates that semantic similarity, as captured by embedding-based methods, is not always the right approach. Skill-targeted adaptive training provides a direct way to  target the weaknesses of the model.

%% file: main_sections/related_work.tex
\section{Discussion}

\textbf{Related Works:} We provide a more detailed discussion of related works in \Cref{app:related_works}. Broadly, prior approaches can be grouped into three directions. First, several skill-aware algorithms improve language models either by designing more targeted inference-time instructions or by generating synthetic data to instill new skills \citep{kaur2024instruct,gandhi2025cognitive,didolkar2024metacognitive}. In contrast, our method adapts training data toward skills that the model continues to struggle with, even after having been extensively trained.

Second, performance-aware adaptation methods adjust training data to improve efficiency and performance \citep{xia2024less,yu2024mates,xie2023data}. However, these techniques largely focus on minimizing validation loss on a target set, which is only an indirect proxy for generation-time errors. Some attempts to remove dependence on explicit validation sets instead optimize implicit properties such as embedding or gradient diversity \citep{jung2025prismatic,wang2024diversity,yu2024diversify,ni2024exploring}. By contrast, our approach explicitly targets the model’s generation mistakes through a metacognitive framework.


Finally, prior works have shown that keeping a teacher in the training loop can be highly effective \citep{zhou2024teaching,gu2024miniplm,zhang2024elad,wang2023scott,zhou2023distillspec,xu2024speculative}.
In these methods, the teacher provides feedback to the student through logits or targeted generations.
In contrast, our skill-aware targeted training offers a simpler and more efficient alternative.
The teacher only identifies missing skills in the student's generations, which are used to create targeted training data.


\textbf{Conclusion:} We investigate whether targeted skill-based training can improve language models when naive re-training yields little benefit. Using a frontier LLM to analyze responses, we construct a skill profile and selectively re-train on relevant examples, achieving significant gains on both in- and out-of-distribution tasks. Ablations show that models often fail on basic skills like algebraic computations, and \ours{} efficiently addresses such gaps with carefully adapting training data.

Our work points to two promising directions for future research. First, since the general skill feedback identified by a frontier LLM can effectively guide student training, it would be valuable to investigate whether these skills correspond to specific mechanistic circuits within the model. Second, while our focus has been on mathematical datasets, exploring whether \ours{} can also improve dimensions such as safety and interpretability presents an interesting avenue for further study.

%% file: main_sections/ethics_statement.tex
\section*{Acknowledgements}

We thank the members of Princeton Language and Intelligence for their valuable discussions and feedback.
We are also grateful to Anirudh Goyal for his insightful guidance and discussions on skill-targeted training. 
Sanjeev Arora acknowledges support from the NSF, DARPA, ONR, and the Schmidt Foundation.
Abhishek Panigrahi acknowledges support from Apple AIML and Siebel Scholarships.

\section*{Ethics Statement}
All authors of this work have read and agree to abide by the ICLR Code of Ethics. We affirm that this research was conducted in compliance with the principles of research integrity, fairness, and transparency outlined therein.

Our study focuses on developing and evaluating a novel fine-tuning approach for language models, targeted at improving mathematical reasoning benchmarks. The work exclusively utilizes publicly available datasets such as MATH, AMC23, and AIME24/25. These datasets are widely used in the research community and do not involve human subjects, private data, or personally identifiable information. No sensitive, proprietary, or confidential data were accessed or released.

We acknowledge that research in language model training can have broader societal impacts, particularly regarding potential misuse (e.g., generating misleading or harmful content). However, our contributions are specifically focused on mathematical problem-solving and skill-targeted fine-tuning, which pose minimal direct risk of harmful applications. The methods proposed are not designed for, nor evaluated on, domains involving sensitive personal, social, or political content.

We have no conflicts of interest or external sponsorships that could bias the reported results. All experiments were performed under standard academic conditions with openly available resources. Our work complies with legal and ethical standards for dataset usage, algorithm development, and reporting.

\textbf{Use of LLM:} We used an LLM solely to improve the clarity and readability of the manuscript text (e.g., grammar and style polishing). The model was not employed for designing experiments, analyzing data, or generating results. All scientific contributions, methodologies, and findings reported in this work are the product of the authors.

\section*{Reproducibility Statement}
We have taken several steps to ensure the reproducibility of our results. A detailed description of the \ours{} algorithms, including pseudocode, is provided in \Cref{sec:design} and \Cref{app:algorithm}. The datasets used in all experiments (MATH, AMC23, AIME24/25, GSM8K, and others) are publicly available and fully cited in the references. We describe our experimental setup, model configurations, training hyperparameters, and ablations in \Cref{sec:expt-set-up} and \Cref{app:expt_details}. 

To facilitate replication, we will release our code repository along with all the \syn{} data if we proceed to the camera-ready version. Together, these resources provide sufficient detail for independent researchers to reproduce our results and extend our methods to related benchmarks.

%% file: appendix/related_work.tex
\section{Related Works} \label{app:related_works}

Recent works show that cognitive theories of human learning can also improve language model performance. \citet{arora2023theory} argue that language models generalize beyond training data by learning transferable skills that connect text tokens. Building on this idea, \citet{wu2024conceptmix,yu2023skill,zhao2024can} propose evaluation benchmarks to test how well models generalize. \citet{didolkar2024metacognitive,he2025adaptmi} use the same framework to design instance-specific in-context examples that improve model's inference-time performance. Closest to our work, \citet{kaur2024instruct} synthesize instruction-following datasets by combining arbitrary skills and show that small models learn more efficiently from such data. Similarly, \citet{gandhi2025cognitive} find that certain cognitive skills are necessary for exploration during reinforcement learning, and these can be encouraged through targeted continual pretraining.
In contrast, we show that we can use the skill-based framework to create targeted training datasets by analyzing the missing skills in model's responses after training and even unlock further gains.

Influence estimation methods have proven effective for constructing targeted training datasets \citep{xia2024less,yeh2022first,kwon2023datainf,penedo2024fineweb,engstrom2024dsdm}. These methods measure the similarity between training and validation data, using gradients or embeddings, to identify the most useful subsets of training examples. In particular, gradient-based approaches estimate how each training example affects the validation loss, then select data with the highest positive influence. However, minimizing validation loss does not always align with improving evaluation metrics, due to the mismatch between average token loss and auto-regressive generation \citep{arora2022exposure,fang2024wrong}. Moreover, such strategies require access to ground-truth solutions, often provided by a strong teacher model, on the validation set.
In contrast, our approach provides a complementary, meta-cognitive alternative. We use a teacher model not to generate ground-truth solutions, but to analyze the student’s responses and identify the missing skills in its generations, directly targeting the model’s weaknesses.

Embedding-based strategies provide an alternative for influence estimation \citep{penedo2024fineweb,li2023textbooks}. However, as shown in our experiments (\cref{sec:expt-set-up}), these methods primarily capture surface-level semantic similarity between the validation and training sets and fail to identify fine-grained weaknesses in model performance. Other works \citep{wang2024diversity,yu2024diversify,ni2024exploring} use embedding-based methods to enhance the diversity of training data. Whether combining such diversity-oriented approaches with our targeted data construction can yield even greater improvements remains an open question for future research.

\looseness-1Finally, we introduce \syn{}, an approach analogous to \sel{}, which synthesizes new training data targeted to the identified missing skills. Synthetic data generation has recently gained attention as a practical way to augment real-world datasets, improving language model performance both in-distribution and out-of-distribution \citep{jung2025prismatic,yu2023metamath,lu2023instag,li2023generaltextembeddingsmultistage,kaur2024instruct}.
Our goal is not to propose the best synthetic data generation method, but to demonstrate the effectiveness of a metacognition-based strategy for creating targeted training data for the student model. A comprehensive comparison of \syn{} with existing synthetic data generation techniques is left for future work.

%% file: appendix/design_details.tex
\section{Details of \ours{} data creation} 
\label{app:data-creation-details}

\subsection{Algorithm for constructing \sel{} and \syn{} data} \label{app:algorithm}
\Cref{alg:selection} outlines the procedure to construct \ptarget{} in Stage 3 (\Cref{sec:design}). For each question in the test set \test{}, the algorithm first identifies the associated missing skills using the Missing-Skill Profile. For each missing skill, a small set of examples is retrieved from the Skill-Map, which links each skill to corresponding training data. In \sel{}, these retrieved examples are directly added to the target training set. Otherwise, the examples are used as seeds to prompt GPT-4o to generate new, skill-specific questions, which are then included instead. This process ensures that the resulting training set \ptarget{} is adaptively enriched with examples that directly address the model’s weaknesses.

\begin{algorithm}[htbp]
\caption{Skill-based data selection/generation}
\label{alg:selection}
\textbf{Input:} Test set \test{}, \texttt{Skill-Map}: \skills{}$\to$\train{}, \texttt{MissingSkillProfile}: \test{}$\to$\skills{}, \sel{}: bool \\
\textbf{Output:} \ptarget{}
\begin{algorithmic}[1]
\State \ptarget{} $\gets$ \texttt{[]}
\For{$q$ in \test{}} 
        \State \texttt{skill\_list} $\gets$ \texttt{MissingSkillProfile[q]}
        \If{\texttt{skill\_list} is not empty}
            \For{\texttt{skill} in \texttt{skill\_list}} 
            \State \train{}$_{skill}$ $\gets$ \texttt{Skill-Map[skill]}
            \State \train{}$_{selected}$ $\gets$ \texttt{random\_sample(\train{}$_{skill}$, 3)}
            \If{\sel{}}
            \State \ptarget{} $\gets$ \ptarget{} + \train{}$_{selected}$
            \Else
            \State \train{}$_{new}$ $\gets$ GPT-4o(\train{}$_{selected}$, \texttt{skill}, prompt="Propose a new question based on \\\hspace{7.5cm} the given questions and the given skill.")
            \State \ptarget{} $\gets$ \ptarget{} + \train{}$_{new}$
            \EndIf

            \EndFor 
        \EndIf
\EndFor 

\State \Return \ptarget{}
\end{algorithmic}
\end{algorithm}

\subsection{Training Data Creation Procedure of \ours{}} \label{app:data-creation-ours}
We now provide a detailed interpretation of our training data creation approach outlined \Cref{alg:selection}.
\paragraph{\sel{}.}
\underline{4k unique questions, 9.5k QA pairs.}
We begin by filtering $500$ difficult questions from the validation set using our process reward model. For each such question, the teacher model identifies $2$–$3$ missing skills in the student’s response. As described in \Cref{sec:stage3}, we then create the training set by selecting $5$ questions for each missing skill in the question’s \profile{}. We use $3$ answers for each question and randomly sample a subset of 9.5k question-answer pairs as our training set.

\paragraph{\syn{}.}
\underline{4k unique questions, 9.5k QA pairs.} 
We begin by filtering $500$ difficult questions from the validation set using our process reward model. For each such question, the teacher model identifies $2$–$3$ missing skills in the student’s response. For each pair of \texttt{(difficult\_question, missing\_skill)}, we retrieve $3$ questions from \math{} training set. We input these $3$ questions, along with the \texttt{missing\_skill}, to the teacher model, prompting it to synthesize $2$ new questions. The teacher further generates $3$ solutions for each new question. We then filter the newly synthesized data by:

\quad \textbf{1.} Compute consistency scores for each set of \texttt{(new\_question, solution)} pairs, according to the number of solutions agreeing on the final answer. For example, a new question with $2$ solutions agreeing on the final answer has a consistency score of $2$.

\quad \textbf{2.} Keep only the \texttt{new\_question} with a consistency score of $\geq2$.

\quad \textbf{3.} For each filtered question, keep only the \texttt{solution} that agrees on the final answer. \footnote{For \syn{}, after filtering teacher-generated answers using consistency, we obtained 9.5k valid question–answer pairs. To ensure comparability, we standardize the training data size to 9.5k pairs for all experiments.}

This process enables our approach to generate diverse data, as we input $3$ questions to the teacher model as references each time. The consistency-filtering step filters out both invalid questions and solutions, ensuring the quality of \syn{}.

%% file: appendix/full_experiment_details.tex
\newpage
\section{Experimental details} \label{app:expt_details}

\subsection{Model \& Training Configurations} \label{app:model-training-config}
\textbf{Model Settings.} All inferences are under 0-shot settings, with temperature 0.1 for pass@1 sampling, and temperature 1.0 for average@64 or pass@64 sampling. 
For the process reward model in Stage 1 (\Cref{sec:stage1}), we use RLHFlow/Llama3.1-8B-PRM-Mistral-Data (\cite{xiong2024rlhflowmath}), an 8B process reward model fine-tuned from Llama-3.1-8B, with filtering thresholds $\tau_1=0.85,\tau_2=0.7$. We use seed=0 for all evaluations.

\textbf{SFT configurations.} For SFT, we adopt QLoRA with rank 16, scaling factor $\alpha$ = 32, and dropout 0.05, applied to the attention and MLP projection modules. Models are trained in 4-bit NF4 quantization with bfloat16 compute, using the paged AdamW (8-bit) optimizer. We train for 3 epochs with a cosine learning rate schedule and a 3\% warmup ratio. Peak learning rate is chosen separately for each method among \{5e-4, 2e-4, 1e-4, 8e-5, 2e-5, based on accuracy on \math{}. The effective batch size is 8 (per-device batch size of 2 with gradient accumulation of 4). We apply gradient clipping at 0.3, weight decay of 0.1, and enable group-by-length packing for efficiency. Other configurations follow the official code base from Llama\footnote{https://github.com/meta-llama/llama-cookbook} and Qwen\footnote{https://github.com/QwenLM/Qwen}.

\textbf{GRPO configuration.} We train for 6 epochs using a constant learning rate of 5e-7. The objective includes only the policy update loss, without any KL-divergence term, and the entropy coefficient is fixed at $0.0$. Each batch contains $256$ questions, with $4$ rollouts generated per question. Responses are truncated at a maximum length of $2048$ tokens. We set the PPO mini-batch size to $64$, which implies that each batch of $256$ questions is split into four mini-batches. The model performs four gradient updates before refreshing the reference model.

\subsection{Training Data Creation Procedure of Baselines} \label{app:data-creation}
We compare \sel{} and \syn{} with the following baseline models fine-tuned with various data selection/generation methods, to measure the effectiveness of skill-aware training:

\quad$\bullet$ \textbf{\naive{}:} \underline{7.5k unique questions, 7.5k QA pairs.} We naively train the model on all question from the training dataset, with a single answer from the original dataset for each question.

\quad$\bullet$ \textbf{\augmented{}:} \underline{7.5k unique questions, 9.5k QA pairs.} In order to make a fair comparison to our proposed methods, we pick $3$ answers per question to create 22.5k question-answer pairs and then randomly sample a subset of 9.5k question answer pairs as our training set.

\quad$\bullet$ \textbf{\hard{}:} \underline{3k unique questions, 9.5k QA pairs.} We include all questions from the  levels 4 and 5 of the MATH dataset. We use $3$ responses per question to create a pool of 12k question-answer pairs and then keep a random subset of 9.5k question answer pairs.

\quad$\bullet$ \textbf{\embedsel{}:} \underline{4k unique questions, 9.5k QA pairs.} Here, we compare the effectiveness of skill-based training data selection to embedding-based training data selection \footnote{We use Alibaba-NLP/gte-Qwen2-7B-instruct as our embedding model \citep{li2023generaltextembeddingsmultistage}}. We use our {\em difficult} question set from stage 1 and for each question, we pick $5$ similar questions from the training set using an embedding model based similarity score. We pick $3$ answers per selected questions and keep a random subset of 9.5k question answer pairs.

\quad$\bullet$ \textbf{\embedsyn{}:} \underline{4k unique questions, 9.5k QA pairs.} For each question in the difficult set identified during stage 1, we retrieve $5$ question–answer pairs from the training set $\mathcal{P}$ using an embedding-based similarity measure. The teacher model is then prompted to generate $5$ new questions, each accompanied by $3$ candidate responses, conditioned on different groups of $3$ retrieved pairs as in-context examples. We retain only those generated questions for which the LLM produces at least 2 consistent responses, and add the corresponding consistent question–answer pairs to our training set. Finally, we keep a random subset of 9.5k question answer pairs to create our training set.

\newpage
\subsection{Prompts} \label{app:prompts}
\subsubsection{Constructing Skill-Map on \math{}}
\label{app: skill-label}
\paragraph{Statistics of skill lists.} We adopt the list of mathematical skills obtained in \cite{didolkar2024metacognitive} using an LLM labeling$\rightarrow$clustering pipeline. The skill list contains 128 skills in total, divided into 7 subsets across 7 subjects. Each subject includes $\sim$$18$ skills.

\paragraph{Skill-Map construction procedure.} To construct the \skillmap{} (see \Cref{sec:design}), we follow \cite{didolkar2024metacognitive} to label skills on both the training and test sets of \math{} using \gpt{} \citep{openai2024gpt4omini}. We enlist all skills that we used to annotate the questions in \math{} and dataset in \Cref{tab:skill_list_1,tab:skill_list}, which have been taken from \cite{didolkar2024metacognitive}.
We ask the LLM to read the question and provide up to five skills required to solve this question, from the given existing skill list.
We show an example prompt for annotating MATH Number Theory questions as follows.

\begin{tcolorbox}[title=Example skill annotation prompt for MATH Number Theory questions]
[TASK]\\
  You'll be given a math question. Your task is to output:\\
  (1) $<$ skill$>$ list here up to five skill(s) that are required to solve this problem, seperated by commas $<$/skill$>$.\\
  (2) $<$reason$>$ reason here why these skills are needed $<$/reason$>$.\\

  [SKILL LIST]\\
  You should only choose the skills from this list:\\
  \text{[} \\
    "arithmetic\_sequences",\\
    "base\_conversion",\\
    "basic\_arithmetic",\\
    "division\_and\_remainders",\\
    "exponentiation",\\
    "factorization",\\
    "greatest\_common\_divisor\_calculations",\\
    "modular\_arithmetic",\\
    "number\_manipulation",\\
    "number\_theory",\\
    "polynomial\_operations",\\
    "prime\_number\_theory",\\
    "sequence\_analysis",\\
    "solving\_equations",\\
    "understanding\_of\_fractions"\\
  \text{]} \\
  
  [QUESTION]\\
  \{question\}\\

  [REASON AND SKILL(S)]\\
\end{tcolorbox}

\Cref{relabeled-skills} shows some example \math{} questions and their corresponding annotated skills. From the skill annotation, we construct a \skillmap{} (see \Cref{sec:design}) that stores the required skills for each question.
\begin{table}[h]
\begin{center}
\begin{tabular}{>{\raggedright\arraybackslash}p{7cm} 
                >{\raggedright\arraybackslash}p{5.5cm}}
\toprule
\multicolumn{1}{c}{Question}  & \multicolumn{1}{c}{Annotated skills} \\
\midrule
What is the units digit of $3^1 + 3^3 + 3^5 + 3^7 + \ldots + 3^{2009}$? &
exponentiation, modular arithmetic, sequence analysis \\\midrule

In the addition problem  each letter represents a distinct digit. What is the numerical value of E? [Figure] &
basic arithmetic, number manipulation, solving equations \\\midrule

In triangle $ABC$, $\tan(\angle CAB)$ = $\frac{22}{7}$, and the altitude from $A$ divides $\overline{BC}$ into segments of length 3 and 17. What is the area of triangle $ABC$? &
geometry and space calculation, trigonometric calculations, arithmetic operations\\
\bottomrule
\end{tabular}
\end{center}
\caption{Example \math~questions, and the annotated skills generated by \gpt{}.}\label{relabeled-skills}
\end{table}

\begin{table}[H]
    \centering
    \renewcommand{\arraystretch}{1.2}
    \begin{tabular}{>{\raggedright\arraybackslash}p{3.5cm} | >{\raggedright\arraybackslash}p{10cm}}
        \toprule
        \textbf{Subject} & \textbf{List of Skills} \\
        \midrule
        \multicolumn{2}{c}{Per subject split in \math{} } \\
        \midrule
        Algebra &  \texttt{algebraic\_expression\_skills}, 
            \texttt{algebraic\_manipulation\_skills}, 
            \texttt{arithmetic\_skills}, 
            \texttt{calculation\_and\_conversion\_skills}, 
            \texttt{combinatorial\_operations\_and\_basic\_arithmetic}, 
            \texttt{complex\_number\_skills}, 
            \texttt{distance\_and\_midpoint\_skills}, 
            \texttt{exponent\_and\_root\_skills}, 
            \texttt{factoring\_skills}, 
            \texttt{function\_composition\_skills}, 
            \texttt{function\_skills}, 
            \texttt{geometric\_sequence\_skills}, 
            \texttt{graph\_and\_geometry\_skills}, 
            \texttt{inequality\_skills}, 
            \texttt{logarithmic\_and\_exponential\_skills}, 
            \texttt{number\_theory\_skills}, 
            \texttt{polynomial\_concepts}, 
            \texttt{quadratic\_equation\_skills}, 
            \texttt{ratio\_and\_proportion\_skills}, 
            \texttt{sequence\_and\_series\_skills}, 
            \texttt{solving\_equations} \\
            \midrule
        Counting and Probability &  \texttt{calculating\_and\_understanding\_combinations}, 
            \texttt{combinatorial\_mathematics}, 
            \texttt{combinatorics\_concepts}, 
            \texttt{counting\_principals}, 
            \texttt{factorials\_and\_prime\_factorization}, 
            \texttt{number\_theory\_and\_arithmetic\_operations}, 
            \texttt{permutation\_and\_combinations}, 
            \texttt{probability\_calculation\_with\_replacement}, 
            \texttt{probability\_concepts\_and\_calculations}, 
            \texttt{probability\_theory\_and\_distribution}, 
            \texttt{combinatorics\_operations} \\
            \midrule
        Geometry & \texttt{3d\_geometry\_and\_volume\_calculation\_skills},  
\texttt{algebraic\_skills},  
\texttt{area\_calculation\_skills},  
\texttt{circle\_geometry\_skills},  
\texttt{combinatorics\_and\_probability\_skills},  
\texttt{coordinate\_geometry\_and\_transformation\_skills},  
\texttt{other\_geometric\_skills},  
\texttt{pythagorean\_skills},  
\texttt{quadrilateral\_and\_polygon\_skills},  
\texttt{ratio\_and\_proportion\_skills},  
\texttt{triangle\_geometry\_skills},  
\texttt{trigonometry\_skills},  
\texttt{understanding\_circle\_properties\_and\_algebraic\_manipulation} \\
        \bottomrule
    \end{tabular}
    \caption{List of skills used for annotating questions in each subject in \math{} dataset}
    \label{tab:skill_list_1}
\end{table}

\begin{table}[H]
    \centering
    \renewcommand{\arraystretch}{1.2}
    \begin{tabular}{>{\raggedright\arraybackslash}p{3.5cm} | >{\raggedright\arraybackslash}p{10cm}}
        \toprule
        \textbf{Subject} & \textbf{List of Skills} \\
        \midrule
        \multicolumn{2}{c}{Per subject split in \math{} } \\
        \midrule
        Intermediate Algebra & \texttt{absolute\_value\_skills},  
\texttt{algebraic\_manipulation\_and\_equations},  
\texttt{calculus\_optimization\_skills},  
\texttt{complex\_number\_manipulation\_and\_operations},  
\texttt{function\_composition\_and\_transformation},  
\texttt{graph\_understanding\_and\_interpretation},  
\texttt{inequality\_solving\_and\_understanding},  
\texttt{polynomial\_concepts},  
\texttt{properties\_and\_application\_of\_exponents},  
\texttt{quadratic\_equations\_and\_solutions},  
\texttt{recursive\_functions\_and\_sequences},  
\texttt{sequence\_and\_series\_analysis\_skills},  
\texttt{simplification\_and\_basic\_operations},  
\texttt{solving\_inequalities},  
\texttt{solving\_system\_of\_equations},  
\texttt{summation\_and\_analysis\_of\_series},  
\texttt{understanding\_and\_application\_of\_functions},  
\texttt{understanding\_and\_applying\_floor\_and\_ceiling\_functions},  
\texttt{understanding\_and\_manipulation\_of\_rational\_functions},  
\texttt{understanding\_and\_utilizing\_infininte\_series},  
\texttt{understanding\_ellipse\_properties},  
\texttt{understanding\_logarithmic\_properties\_and\_solving\_equations}\\
\midrule
        Number Theory & \texttt{arithmetic\_sequences},  
\texttt{base\_conversion},  
\texttt{basic\_arithmetic},  
\texttt{division\_and\_remainders},  
\texttt{exponentiation},  
\texttt{factorization},  
\texttt{greatest\_common\_divisor\_calculations},  
\texttt{modular\_arithmetic},  
\texttt{number\_manipulation},  
\texttt{number\_theory},  
\texttt{polynomial\_operations},  
\texttt{prime\_number\_theory},  
\texttt{sequence\_analysis},  
\texttt{solving\_equations},  
\texttt{understanding\_of\_fractions}
\\
\midrule
        Pre-algebra & \texttt{average\_calculations},  
\texttt{basic\_arithmetic\_operations},  
\texttt{circles},  
\texttt{counting\_and\_number\_theory},  
\texttt{exponentiation\_rules},  
\texttt{fractions\_and\_decimals},  
\texttt{geometry},  
\texttt{multiples\_and\_zero\_properties},  
\texttt{multiplication\_and\_division},  
\texttt{perimeter\_and\_area},  
\texttt{prime\_number\_theory},  
\texttt{probability\_and\_combinatorics},  
\texttt{ratio\_and\_proportion},  
\texttt{linear\_equation}
\\
\midrule
        Pre-calculus & \texttt{algebra\_and\_equations},  
\texttt{basic\_trigonometry},  
\texttt{calculus},  
\texttt{complex\_number\_operations},  
\texttt{complex\_numbers},  
\texttt{coordinate\_systems},  
\texttt{determinant\_calculation},  
\texttt{geometric\_relations},  
\texttt{geometry\_and\_space\_calculation},  
\texttt{geometry\_triangle\_properties},  
\texttt{matrix\_operations},  
\texttt{parametric\_equations},  
\texttt{sequences\_series\_and\_summation},  
\texttt{three\_dimensional\_geometry},  
\texttt{trigonometric\_calculations},  
\texttt{vector\_operations}
\\
        \bottomrule
    \end{tabular}
    \caption{List of skills used for annotating questions in each subject of \math{}  dataset (continued from \Cref{tab:skill_list_1})}
    \label{tab:skill_list}
\end{table}

\newpage

\subsubsection{Missing skill Identification from Model Responses}
In Stage 2 of \ours{}, we use \gpt{} to label the skills that are missing from a model response. We ask the LLM to read the question along with the SLM response and provide the skills that the model fails to leverage in the response, from the given existing skill list.
Below we show an example prompt for labeling missing skills for MATH Number Theory questions, as well as an example LLM output.

\begin{tcolorbox}[title=Example missing skill identification prompt for a MATH Number Theory question] \label{app: missing-skill-prompt}
[TASK]{\\}
   You'll be given a math question and a step-by-step solution written by a Small Language Model. Your task is to output:{\\}
  (1) $<$judge$>$ judge here whether the solution is correct or incorrect  $<$/judge$>${\\}
  (2)  $<$reason$>$ if it's incorrect, reason here why the solution is incorrect  $<$/reason$>$,{\\}
  (3)  $<$skill$>$ list here what skill(s) should the SLM enhance in order to answer correctly, seperated by commas  $<$/skill$>$.{\\}

  [SKILL LIST]{\\}
  \{list of 15 number theory skills\} {\\}
  
  [QUESTION]{\\}
  \{question\} {\\}

  [MODEL\_SOLUTION]{\\}
  \{model\_response\} {\\}

  [YOUR REASON AND SKILL(S)]\\
\end{tcolorbox}
\begin{tcolorbox}[title=Example \gpt{} output for missing skill identification] \label{app: missing-skill-output}
(1) $<$judge$>$ incorrect $<$/judge$>${\\}
(2) $<$reason$>$ The solution incorrectly simplifies the expression and misapplies properties of modular arithmetic. The solution claims that the sum of the inverses from 1 to ( p-1 ) is (-1 pmod{p}), which is incorrect. The correct approach involves recognizing that each term $( a^{-1} \cdot (a+1)^{-1} )$ should be evaluated individually, and the sum should be calculated directly without assuming incorrect simplifications. The solution also fails to correctly apply the properties of modular inverses and does not properly evaluate the sum. $<$/reason$>${\\}
(3) $<$skill$>$ modular\_arithmetic, number\_theory, understanding\_of\_fractions $<$/skill$>$
\end{tcolorbox}

\newpage
\subsubsection{Proposing new questions from skill-based training question distribution}
\label{app:propose-question}
In Stage 3 of \syn{}, we use GPT-4o to propose new questions based on existing questions in the skill-based training question distribution. Below we show an example prompt.

\begin{tcolorbox}[title=Example prompt for proposing new questions using GPT-4o] \label{app: proposing-prompt}
[TASK]{\\}
You'll be given three math questions (e.g., [QUESTION 1]), with their solutions for reference. Your task is to output a new, novel math question that emphasizes the use of [SKILL].{\\}
Important Note: the new question should not be very similar to any of the given questions (e.g., naive adaptions like altering variable names or values from a given question is strictly prohibitted). Meanwhile, the new question should have similar difficulty with the given questions.{\\}
Output format:{\\}
(1) $<$reason$>$ reason here how the given questions relates to the [SKILL] $<$/reason$>${\\}
(2) $<$draft$>$ reason here how to design a new, novel question while emphasizing the [SKILL] $<$/draft$>${\\}
(3) $<$question$>$ your newly constructed math question $<$/question$>${\\}

[QUESTION 1]{\\}
\{train\_set\_question1\}{\\}

[QUESTION 2]{\\}
\{train\_set\_question2\}{\\}

[QUESTION 3]{\\}
\{train\_set\_question3\}{\\}

[SKILL]{\\}
\{missing\_skill\}{\\}
\end{tcolorbox}

%% file: appendix/additional_results.tex
\newpage
\section{Additional Results} \label{app:additional-results}

\subsection{Evaluation results on \llamaS{}}
\label{app:evaluation-results}
\Cref{tab:llamaS-results} shows the evaluation results on \llamaS{}. Similar to \Cref{tab:main-results}, \ours{}
consistently outperforms both heuristic-based and embedding-based data augmentation baselines on in-distribution dataset and most OOD benchmarks. We presented more discussion in \Cref{sec:results} and \Cref{sec:discussion}.

\begin{table*}[htbp]
    \centering
    \small
    \fontsize{8.5pt}{10.5pt}\selectfont
    \setlength{\tabcolsep}{4.7pt}
    \begin{tabular}{lcc@{\hspace{2.3pt}}c@{\hspace{2.3pt}}c@{\hspace{3pt}}cccccc}
        \toprule
         \multirow{2}{*}{Models} 
         & \multirow{2}{*}{\textbf{MATH}} 
         & \multirow{1.9}{*}{\textbf{MATH$^{\mathbf{D}}$}} 
         & \multirow{1.9}{*}{\textbf{MATH$^{\mathbf{2}}$}}  
         & \multirow{2}{*}{\textbf{GSM8K}} 
         & \multirow{2}{*}{\textbf{AMC23}} 
         & \multicolumn{2}{c}{\textbf{MATH-perturb}} 
         & \multicolumn{2}{c}{\textbf{AIME}}
         & \multirow{2}{*}{\textbf{Avg.}} 
         \\
         \cmidrule(lr){7-8} \cmidrule(lr){9-10}
        & & & & & & \textbf{simple}& \textbf{hard} & \textbf{2024}  &\textbf{2025}\\
        \midrule

\rowcolor{sage1}
    \multicolumn{11}{c}{\textit{\llamaS{} $+$ SFT}}\\

Base Model & 26.0 & 15.1 & 9.1 & 40.7 & \underline{11.1} & 17.2 & 6.5 & 20.0 & \underline{10.0}& 17.3\\

\naive{}  & 27.0 & 14.5 & 10.0 & 42.8 & 8.8 & 19.0 & 6.8 & \textbf{26.7} & \underline{10.0}& 18.4\\
\augmented{}  & 27.8 & 14.2 & 8.1 & 43.4 & \underline{11.1} & 17.9 & 6.8 & \textbf{26.7} & 3.3& 17.7\\
\hard{}  & 28.4 & 15.4 & 8.6 & 44.6 & 10.8 & 18.6 & 7.2 & \underline{23.3} & 3.3& 17.8\\
\embedsel{} & 27.4 & 15.6 & 8.6 & 44.6 & 8.8 & 18.6 & 6.8 & \textbf{26.7} & 3.3 & 17.8\\

\embedsyn{}  & 28.4 & \underline{17.2} & \underline{11.0} & 44.3 & 10.0 & \underline{20.1} & \textbf{7.9} & \underline{23.3} & 6.7 & 18.8\\

\textbf{\sel} & \underline{32.4} & 15.6 & \underline{11.0} & \underline{45.0} & \textbf{12.0} & 19.4 & \textbf{7.9} & \textbf{26.7} & \textbf{16.7}& {\cellcolor{sage2}\textbf{20.7}}\\

\textbf{\syn}  & \textbf{34.5} & \textbf{18.3} & \textbf{12.4} & \textbf{45.6} & 11.0 & \textbf{20.8} & 7.5 & \underline{23.3} & \underline{10.0}& {\cellcolor{pinkintro}\underline{20.4}}\\

\rowcolor{lightgray}
\multicolumn{11}{c}{\textit{$+$ GRPO}}\\
Base Model & 31.8 & 14.4 & 9.5 & 49.7 & 13.3 & 23.3 & \underline{8.2} & 20.0 & 6.7 & 19.7\\

{\naive{}} & 32.0 & 16.0 & \underline{11.9} & \underline{50.8} & 10.0 & \underline{23.7} & 7.9 & 16.7 & 6.7 & 19.5\\

{\augmented{}} & 31.2 & 15.0 & 9.0 & 49.1 & 13.6 & \textbf{24.7} & 7.9 & 23.3 & \underline{13.3} & 20.8\\

{\hard{}} & 32.2 & 14.8 & 11.0 & 50.6 & 11.6 & 22.9 & 6.5 & 26.7 & 10.0 & 20.7\\

{\embedsel{}} & 32.8 & 16.2 & 11.4 & 49.9 & 12.0 & 21.9 & 6.5 & 23.3 & \underline{13.3} & 20.8\\

{\embedsyn{}} & 32.6 & 15.0 & 10.5 & \textbf{51.0} & \underline{13.9} & 21.1 & 6.8 & 26.7 & 3.3 & 20.1\\

{\textbf{\sel{}}} & \underline{34.8} & \underline{16.6} & \textbf{13.8} & 50.1 & \textbf{14.8} & \underline{23.7} & \textbf{9.0} & \underline{30.0} & \textbf{16.7} & {\cellcolor{sage2}\underline{23.3}}\\

{\textbf{\syn{}}} & \textbf{35.2} & \textbf{21.1} & \textbf{13.8} & \textbf{51.0} & \textbf{14.8} & \textbf{24.7} & 7.9 & \textbf{33.3} & \textbf{16.7} & {\cellcolor{pinkintro} \textbf{24.3}}\\
    \bottomrule
\end{tabular}
\caption{Improvements on various math benchmarks from applying STAT.
    Results under ‘+SFT’ show the performance of SFT models trained with each method, while ‘+GRPO’ shows the performance after applying GRPO on top of the corresponding SFT models. Our methods, \sel{} and \syn{}, achieve an average gain of up to 3.4\% over the base model, with strong OOD performances (AMC23 results reported on average@64, AIME on pass@64). Applying GRPO on top of fine-tuning with \ours{} further enhances these improvements. See \Cref{tab:main-results} for results on \llamaM{} and \qwM{}.}
    \label{tab:llamaS-results}
\end{table*}

\subsection{Missing-Skill-Profile}
\label{app:missing-skill-profile}

\begin{figure}[h!]
    \centering
    \includegraphics[width=\linewidth]{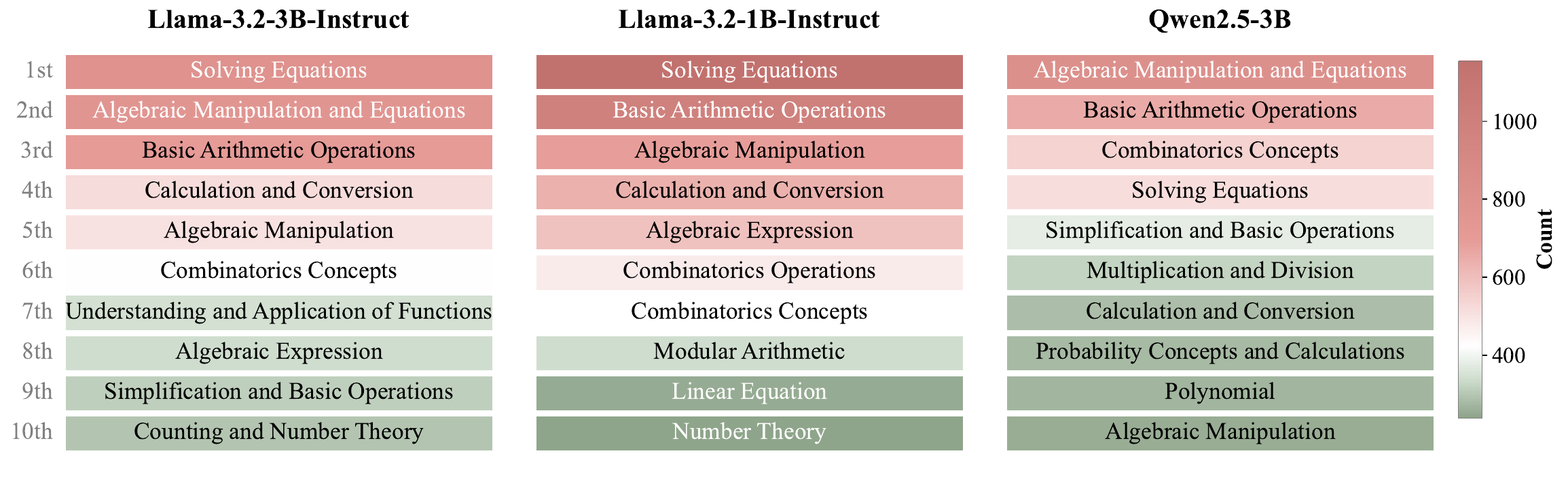}
    \vspace{-10pt}
    \caption{Top 10 missing skills of \llamaM{}, \llamaS{}, and \qwM{}. The models struggle most with fundamental mathematical skills such as solving equations and basic arithmetic operations.} 
    \label{fig:skill-profile}
\end{figure}

\Cref{fig:skill-profile} shows the snippets of model-specific $\text{Missing-Skill-Profile}$ of \llamaM{}, \llamaS{}, and \qwM{}, obtained at the end of Stage 2 (see \Cref{sec:stage2}). These profile snippets include the Top 10 frequent missing skills of the models. As discussed in \Cref{sec:discussion}, most of the frequent missing skills in both models are algebra-related, such as solving equations, manipulation, and calculation.
In addition, both models also demonstrate noticeable weaknesses in conceptual and reasoning-oriented mathematical skills, including combinatorics, understanding and application of functions, and number theory. 
Compared to \llamaM{}, the missing skill profile of \llamaS{} concentrated more towards basic operations (e.g., solving equations), suggesting that smaller models have more pronounced limitations in fundamental computational abilities.

\newpage
\subsection{Case study: example questions with different skills} \label{app:question-case-study}
We observe in \Cref{sec:discussion} that the baseline training data distribution largely misaligns with model's actual missing skills, with baseline data emphasizing more advanced and conceptual skills (e.g., Prime Number Theory, Polynomial Concepts) while the model lacking more basic operational skills (e.g., Solving Equations). To better visualize this misalignment in data distribution, here we showcase three example QA pairs from the \math{} training set, respectively associated with the three skills above.

\begin{tcolorbox}[title=Question with skill ``Solving Equations", label={box:polynomial}] 
\textbf{Question:} The inverse of $f(x) = \frac{2x-1}{x+5}$ may be written in the form $f^{-1}(x)=\frac{ax+b}{cx+d}$, where $a$, $b$, $c$, and $d$ are real numbers.  Find $a/c$. {\\}

\textbf{Solution:} {\\}
If we substitute $f^{-1}(x)$ into our expression for $f$, we get 
\[
f(f^{-1}(x)) = \frac{2f^{-1}(x)-1}{f^{-1}(x)+5}.
\]
Since $f^{-1}(f(x))=x$, we obtain
\begin{align*}
\frac{2f^{-1}(x)-1}{f^{-1}(x)+5} &= x \\
\Rightarrow \quad 2f^{-1}(x)-1 &= x\bigl(f^{-1}(x)+5\bigr) \\
\Rightarrow \quad 2f^{-1}(x)-1 &= x f^{-1}(x)+5x.
\end{align*}

Moving the terms involving $f^{-1}(x)$ to the left-hand side and the remaining terms to the right-hand side, we get
\begin{align*}
2f^{-1}(x)-x f^{-1}(x) &= 5x+1 \\
\Rightarrow \quad f^{-1}(x)(2-x) &= 5x+1 \\
\Rightarrow \quad f^{-1}(x) &= \frac{5x+1}{-x+2}.
\end{align*}

Now we can see that $(a,b,c,d) = (5,1,-1,2)$ for this representation of $f^{-1}(x)$, so
\[
\frac{a}{c} = \frac{5}{-1} = \boxed{-5}.
\]

\end{tcolorbox}

\begin{tcolorbox}[title=Question with skill ``Prime Number Theory", label={box:prime-number-theory}] 
\textbf{Question:}  The positive integers $A,$ $B,$ $A-B,$ and $A+B$ are all prime numbers. The sum of these four primes is{\\}$\bullet$ A. even{\\}$\bullet$ B. divisible by $3${\\}$\bullet$ C. divisible by $5${\\}$\bullet$ D. divisible by $7${\\}$\bullet$ E. prime{\\}Express your answer using a letter, as A, B, C, D, or E.{\\}

\textbf{Solution:} The numbers $A-B$ and $A+B$ are both odd or both even. However, they are also both prime, so they must both be odd. Therefore, one of $A$ and $B$ is odd and the other even. Because $A$ is a prime between $A-B$ and $A+B,$ $A$ must be the odd prime. Therefore, $B=2,$ the only even prime. So $A-2,$ $A,$ and $A+2$ are consecutive odd primes and thus must be $3,$ $5,$ and $7.$ The sum of the four primes $2,$ $3,$ $5,$ and $7$ is the prime number $17,$ so the correct answer is $\boxed{\text{(E)},}$ prime.

\end{tcolorbox}

\begin{tcolorbox}[title=Question with skill ``Polynomial Concepts", label={box:solving-equations}] 
\textbf{Question:}  The polynomial $P(x) = 2x^3 + ax^2 + bx + c$ has the property that the mean of its zeros, the product of its zeros, and the sum of the coefficients are all equal. The $y$-intercept of the graph of $y = P(x)$ is 8. What is $b$?{\\}

\textbf{Solution:} The $y$-intercept of the graph is the point at which $x=0$. At that point, $P(x)=c$, which we are told is equal to 8. Thus, $c=8$. The product of the roots of the given polynomial is $-\frac{c}{2}=-4$. The problem states that the mean of the zeros must also equal $-4$, so the sum of the three zeros (this is a cubic equation) is equal to $3 \cdot -4 = -12$. The sum of the zeros is also equal to $-\frac{a}{2}$, so $a=24$. Finally, we are given that the sum of the coefficients, or $2+ a+b+c$, is also equal to $-4$. Plugging in our known values of $a$ and $c$, we have $2+24+b+8=-4$. Solving for $b$, we get $b=\boxed{-38}$.

\end{tcolorbox}

\newpage
\subsection{Effectiveness of \ours{} on each subject} \label{app:subject-improve}
\begin{figure}[htbp]
    \centering
    \includegraphics[width=0.7\linewidth]{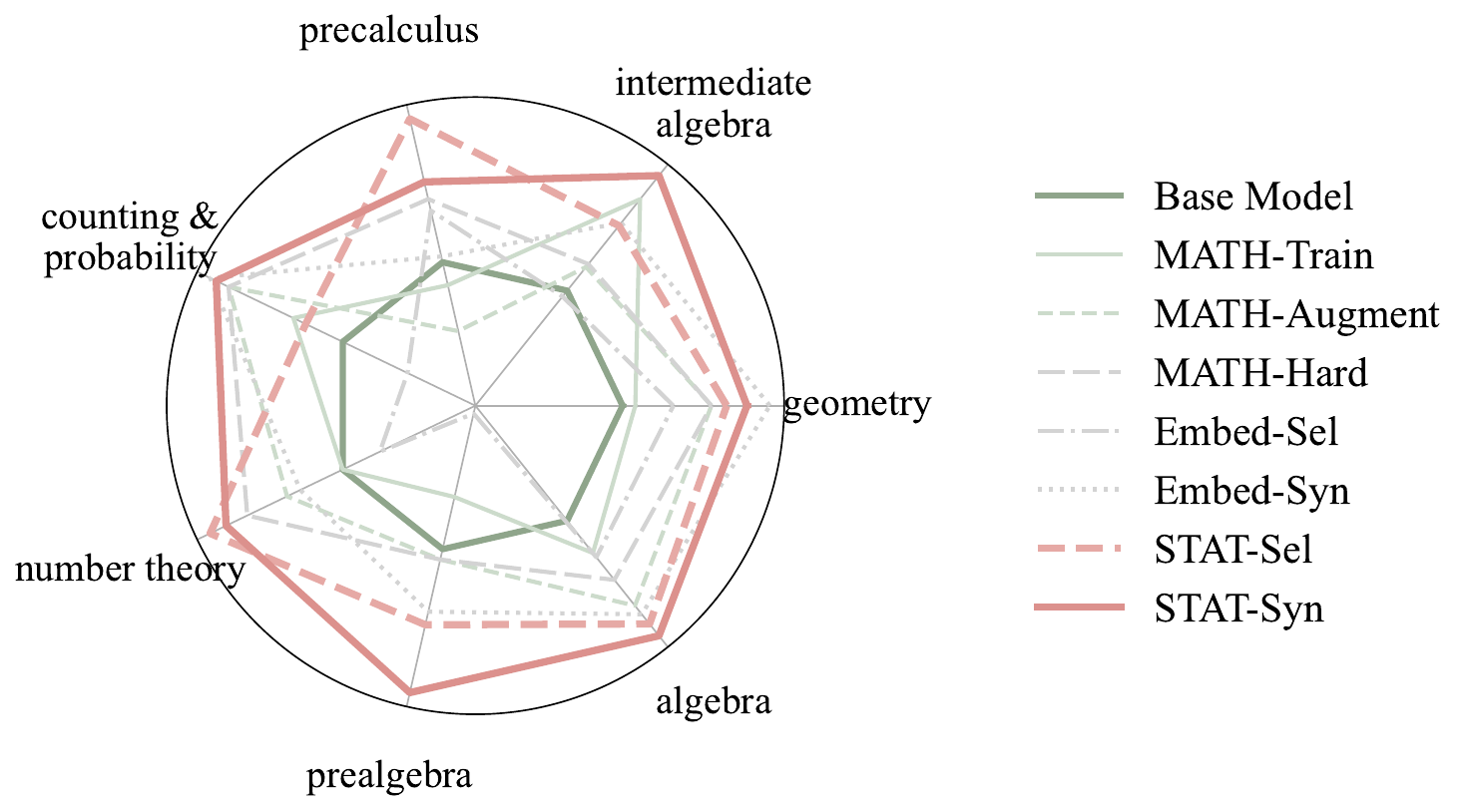}
    \caption{Fine-tuned model performances on \math{} subjects, across different training methods. For better visualization, accuracies are normalized per skill axis, with the base model drawn as a uniform circle and the highest-performing method on each skill placed at the outer edge. \syn{} and \sel{} are most effective in enhancing model performance across nearly all the subjects.} 
    \label{fig:subject-improve}
\end{figure}

To evaluate whether \ours{} enhances general subject-level competence, we measure model accuracy across the 7 subject categories in \math{}. These subjects are: prealgebra, algebra, intermediate algebra, geometry, precalculus, number theory, and counting \& probability. As shown in \Cref{fig:subject-improve}, both \sel{} and \syn{} consistently outperform the base model and data augmentation baselines across nearly all subjects. Notably, \sel{} achieves the strongest improvements in precalculus and number theory, while \syn{} excels in intermediate algebra, prealgebra, algebra, geometry and counting \& probability. It is worth noting that \ours{} brought most improvements on the 3 algebra-related subjects. This aligns with our observation in \Cref{sec:discussion} that \llamaS{} shows its most pronounced weaknesses in algebra, and confirms that our approaches effectively target the skills the model fundamentally lacks.

%% file: appendix/ablation_analysis.tex
\section{Ablation \& Analysis}
\subsection{Ablations on the reward filtering method in Stage 1} \label{app:prm-ablations}
Recall that in Stage 1 of the \ours{} pipeline, we use an off-the-shelf process reward model (RLHFlow/Llama3.1-8B-PRM-Mistral-Data) to score small language models' responses, in order to filter out a set of \textit{difficult} questions for each model. Here, we conduct various ablation studies on the reward filtering process.
\label{ablate_reward}

\paragraph{Effect of threshold values on the reward model prediction.}
We investigated the effect of $\tau_1$ and $\tau_2$ (defined in \Cref{sec:stage1}) on the classification performance of {\em difficult} questions. Specifically, we measure whether our classification of questions as {\em difficult} also corresponds to the correctness of responses assessed using ground-truth labels. In \Cref{tab:RM}, we report four metrics (accuracy / precision / recall / F1) evaluating the prediction accuracy resulting from different filtering thresholds. Note that $\tau_1=0$ or $\tau_2 = 0$ means completely removing the constraints of $\tau_1$ or $\tau_2$. Across all evaluated combinations of threshold values, our choice of the threshold values ($\tau_1=0.85,\tau_2=0.7$) gives a good combination of prediction scores. To further visualize this effect, we conduct \ours{} on top of all combinations of thresholds, and report the final accuracy in \Cref{tab:RM-accuracy}. Our choice of threshold values yields the highest final accuracy among all the combinations.

\begin{table}[h]
\centering
\small
\begin{tabular}{lcccc}
\toprule
$\tau_1 \backslash \tau_2$ & $\tau_2 = 0$ & $\tau_2 = 0.6$ & $\tau_2 = 0.7$ & $\tau_2 = 0.8$ \\
\midrule
$\tau_1 = 0$     & 53 / 0 / 0 / 0         & 80 / 78 / 79 / 79     & 80 / 74 / 88 / 79     & 75 / 66 / 95 / 78 \\
$\tau_1 = 0.8$   & 80 / 79 / 78 / 79      & 80 / 76 / 85 / 80     & 79 / 72 / 90 / 80     & 75 / 66 / 96 / 78 \\
$\tau_1 = 0.85$  & 79 / 74 / 88 / 80      & 79 / 72 / 90 / 80     & \textbf{78} / \textbf{70} / \textbf{92} / \textbf{80}     & 74 / 65 / 96 / 78 \\
$\tau_1 = 0.9$   & 73 / 64 / 95 / 77      & 73 / 64 / 95 / 77     & 72 / 64 / 96 / 77     & 70 / 62 / 97 / 75 \\
\bottomrule
\end{tabular}
\caption{Reward model performance (accuracy / precision / recall / F1) on classifying correct/incorrect responses from \qwS{} on \math{}, accross different thresholds. $\tau_1=0$ or $\tau_2 = 0$ means completely removing $\tau_1$ or $\tau_2$. Our choice of threshold values ($\tau_1=0.85,\tau_2=0.7$) gives a good combination of prediction scores.}
\label{tab:RM}
\end{table}

\begin{table}[h]
\centering
\small
\begin{tabular}{lcccc}
\toprule
$\tau_1 \backslash \tau_2$ & $\tau_2 = 0$ & $\tau_2 = 0.6$ & $\tau_2 = 0.7$ & $\tau_2 = 0.8$ \\
\midrule
$\tau_1 = 0$     & 52.8 & 55.7 & 55.9 & 55.7 \\
$\tau_1 = 0.8$   & 55.1 & 56.3 & 56.2 & 55.6 \\
$\tau_1 = 0.85$  & 55.3 & 56.4 & \textbf{56.4} & 55.6 \\
$\tau_1 = 0.9$   & 55.7 & 55.7 & 55.6 & 55.2 \\
\bottomrule
\end{tabular}
\caption{Final \ours{} performance of  \qwS{} on \math{}, with different thresholds. Our choice of threshold values ($\tau_1=0.85,\tau_2=0.7$) leads to the highest accuracy.}
\label{tab:RM-accuracy}
\end{table}

\paragraph{Out-of-distribution (OOD) prediction performance of reward model.}
Although we primarily evaluated \ours{} on \math{} and \gsm{}, our method can potentially be extended to other math datasets. While the reward model we used in Stage 1 was only trained on the \math{} and \gsm{} distribution, we show that it is capable of scoring responses for various OOD math datasets. \Cref{tab:RM-OOD} reports the reward model's performance on classifying correct/incorrect responses from \qwM{} on four popular math benchmarks: AMC23, AIME24, AIME25, and MATH$^2$. The reward model achieves comparably high performance on scoring model responses on these OOD, significantly more difficult benchmarks, indicating that the model is highly generalizable. This implies the potential to extend our method to new datasets without the need to train a specialized reward model for each one.

\begin{table}[h]
\centering
\small
\begin{tabular}{lcccc}
\toprule
Metric & AMC23 & AIME24 & AIME25 & MATH$^2$ \\
\midrule
Accuracy  & 92.5 & 86.7 & 86.7 & 84.8 \\
Precision & 90.9 & 92.6 & 86.7 & 95.2 \\
Recall    & 95.2 & 92.6 & 100.0 & 88.5 \\
F1        & 93.0 & 92.6 & 92.9 & 91.0 \\
\bottomrule
\end{tabular}
\caption{Reward model prediction metrics across four OOD math benchmarks. Despite not being trained on these benchmarks, the reward model's prediction capability is largely generalizable to them.}
\label{tab:RM-OOD}
\end{table}

\paragraph{Reward Filtering vs. Simple Heuristics for classifying difficult questions.}
Considering the computational overhead of calling a separate PRM, we explored alternative approaches to classifying questions that rely on computation-free simple heuristics. Specifically, we experimented with two heuristic strategies: 
\begin{itemize}
    \item \textbf{Consistency heuristic:} We measure the consistency of the model across five sampled generations per question and classify questions with lower consistency as difficult. Specifically, a question is \textit{difficult} if, among 5 sampled generations, the most common response appears $<$ 2 times.
    \item \textbf{Length heuristic:} We use the length of the model’s responses as a proxy and classify questions with longer responses as difficult. Specifically, a question is \textit{difficult} if the average model response length on this question is $\geq$ 800 words.
\end{itemize}
\Cref{tab:RM-heuristics} shows that both heuristics yield reasonably accurate predictions. Moreover, applying \ours{} on top of these heuristic-classified difficult questions can improve the final accuracy by 2\%. However, we leave a more thorough investigation into the robustness and generalizability of these strategies in relation to PRM-based classification for future work.

\begin{table}[h]
\centering
\small
\begin{tabular}{lcc}
\toprule
\textbf{Classification method} & \textbf{Classification accuracy} \\
\midrule
Consistency Heuristic &  79.8\% \\
Length Heuristic     & 74.2\% \\
Reward Filtering     & 78.0\% \\
\bottomrule 
\end{tabular}
\caption{Performance of consistency heuristic and length heuristic on classifying difficult questions. The classification accuracy of simple heuristics are on par with the reward filtering method.}
\label{tab:RM-heuristics}
\end{table}

\paragraph{Process Reward vs. Outcome Reward.} We also compare the prediction accuracy of our process reward model (PRM) with threshold filtering (see \Cref{sec:stage1}) against directly loading the reward model as an outcome reward model (ORM). Our preliminary experiments indicated $0.9$ as the optimal threshold for the outcome rewards. With $\tau=0.9$, the prediction metrics of the ORM are: 
Precision $= 0.54$ / Recall $= 0.90$ / F1 $= 0.68$, whereas the prediction metrics of the PRM with optimal thresholds are Precision $= 0.70$ / Recall $= 0.92$ / F1 $= 0.80$. Therefore, our method using PRM with threshold filtering is superior to directly using ORM.

\subsection{Statistics of difficult questions}
In Stage 1 of \ours{} (see \Cref{sec:stage1}), we identify a set of \textit{difficult} questions for each individual model using a process reward model along with a filtering heuristic. \Cref{tab:proportion-difficult} reports the proportions of difficult questions classified for different models in each math domain. Compared to \Cref{tab:main-results}, the proportions of difficult questions closely correspond to the accuracy numbers of each model, even though we did not access the ground truth in the whole pipeline. Notably, our classification method captures not only questions that the model gets wrong, but also questions that the model passes with a flawed solution process.

\begin{table}[h]
\centering
\scriptsize
\begin{tabular}{lcccccc}
\toprule
\textbf{Model} & \textbf{Geometry} & \textbf{Precalculus} & \textbf{Algebra} & \textbf{Prealgebra} & \textbf{Intermediate Algebra} \\
\midrule
\qwM{}     & 61.8 & 70.1 & 29.7 & 33.2 & 75.9 \\
\llamaS{}  & 93.5 & 92.0 & 91.4 & 89.7 & 99.0 \\
\llamaM{}  & 68.2 & 82.7 & 45.5 & 48.9 & 85.7 \\
\midrule
\textbf{Model} & \textbf{Count.\&Prob.} & \textbf{Number Theory} & \textbf{MATH Avg.} & \\
\midrule
\qwM{}     & 62.2 & 56.1 & 52.1 \\
\llamaS{}  & 97.9 & 95.2 & 94.0 \\
\llamaM{}  & 65.2 & 62.3 & 62.3 \\
\bottomrule

\end{tabular}
\caption{Proportions of difficult questions (\%) classified by \ours{} for each model. Although our method did not access the ground truth, the proportion of classified difficult questions still closely mirrors each model's accuracy (see \Cref{tab:main-results}) in each domain.}
\label{tab:proportion-difficult}
\end{table}

\subsection{Analysis of the teacher model} \label{app:teacher-analysis}
\paragraph{Teacher model need not be overwhelmingly stronger than student.}
One feature of \ours{} is the demand of a substantially stronger teacher model to supervise the student. In this section, we evaluate this demand by directly comparing teacher and student performances on math reasoning benchmarks. Due to resource constraints, our evaluation is limited to a representative set of benchmarks, but the results are sufficient to illustrate the key trend: the teacher is not strictly dominant, and the student can approach or even match the teacher’s performance within a manageable gap.

As shown in \Cref{tab:compare-teacher-student}, although teacher models obtain higher absolute scores, they are not overwhelmingly stronger than the students. In particular, the gap between \gpt{} and \qwM{} is only around 10 points across \gsm{} and \math{}, a margin that is significant but manageable. This suggests that \ours{} does not strictly rely on a much stronger teacher to succeed. Instead, even when teacher and student are relatively close in ability, the student can still benefit and recover most of the teacher’s performance. This opens up the possibility of self-improvement, where a model iteratively teaches and refines itself without requiring access to an external teacher that is substantially stronger.

\begin{table}[h]
    \centering
    \scriptsize
    \begin{tabular}{lccccc}
    \toprule
     \multirow{2}{*}{Benchmark}&\multicolumn{2}{c}{\textbf{Teacher}}   &  \multicolumn{3}{c}{\textbf{Student}}\\
     \cmidrule(lr){2-3} \cmidrule(lr){4-6}
      &GPT-4o & \gpt{} & \qwM{}  & \llamaM{} & \llamaS{} \\
      \midrule
    \gsm{}&97.0 & 94.0 &80.9&73.0& 40.7\\
    \math{}&73.0 & 69.1 &55.8&44.0&26.0 \\
      MATH-perturb-simple&62.0 & N/A & 43.7 & 33.7 & 17.2\\
      \phard{}&39.4 & N/A & 24.0 & 12.2 & 6.5 \\    
      \bottomrule
    \end{tabular}
    \caption{\textbf{Math reasoning accuracy (\%).} Comparison between teacher models (GPT-4o, \gpt{}) and student models (\qwM{}, \llamaM{}, \llamaS{}) on \gsm{}, \math{}, MATH-perturb-simple, and \phard{}.}
    \label{tab:compare-teacher-student}
\end{table}

\paragraph{Agreement across different teacher models.}
Since our approach relies on a frontier LLM as teacher, a natural concern is potential bias in the missing-skill labeling process. In light of this, we present a preliminary investigation into the level of agreement among different LLMs in missing skill labeling, using an LLM-as-a-judge approach. We first evaluate \gpt{}’s ability to self-verify the correctness of its own predicted missing skills and find that it judges its predictions to be correct 70\% of the time. To further assess the reliability of these predictions, we compute the agreement between \gpt{} and Claude-3.5-Sonnet. The models agree on 43\% of the predicted skills, where agreement is defined as the average fraction of overlapping skills relative to the total number of skills predicted by \gpt{}. Given the fine-grained nature of our skill list, we consider this level of agreement significant.